\documentclass[10pt,twocolumn,letterpaper]{article}

\usepackage{cvpr}
\usepackage{times}
\usepackage{epsfig}
\usepackage{graphicx}
\usepackage{amsmath}
\usepackage{amssymb}
\usepackage[skip=2pt,font=scriptsize,]{subcaption}
\usepackage{multirow}
\usepackage[table]{xcolor}

\cvprfinalcopy 

\def\cvprPaperID{****} 

\begin{document}

\title{A Human and Group Behaviour Simulation Evaluation Framework utilising Composition and Video Analysis}

\author{Rob Dupre\\
Kingston University\\
{\tt\small R.Dupre@kingston.ac.uk}
\and
Vasileios Argyriou \\
Kingston University\\
{\tt\small Vasileios.Argyriou@kingston.ac.uk}
}

\maketitle
\begin{abstract}
   In this work we present the modular Crowd Simulation Evaluation through Composition framework (CSEC) which provides a quantitative comparison between different pedestrian and crowd simulation approaches. Evaluation is made based on the comparison of source footage against synthetic video created through novel composition techniques. The proposed framework seeks to reduce the complexity of simulation evaluation and provide a platform from which the comparison of differing simulation algorithms as well as parametric tuning can be conducted to improve simulation accuracy or providing measures of similarity between crowd simulation algorithms and source data. Through the use of features designed to mimic the Human Visual System (HVS), specific simulation properties can be evaluated relative to sample footage. Validation was performed on a number of popular crowd datasets and through comparisons of multiple pedestrian and crowd simulation algorithms.
\end{abstract}

\section*{Introduction}

Pedestrian and crowd simulation has applications in a wide range of industries including pedestrian facility suitability and capacity \cite{Asano2009}, computer graphics and gaming \cite{Kim2012}, the social sciences \cite{Klugl2009} and engineering \cite{Xi2011}. This broad range of uses has led to extensive research into how crowds and pedestrians move around and interact with their environment.

The problem of how to evaluate these simulation algorithms is a developing area of research. One of the most prominent issues with crowd and pedestrian simulation research is the lack of a simple and suitable form of comparison between different simulation and modelling approaches. This often means that a given methodology is developed and evaluated for a specific purpose, with its wider abilities and properties left unconfirmed. This task is made more difficult as the developed approaches cover a huge range of applications, where evaluation techniques for one are not always applicable to the others.

Generally the evaluation techniques utilised can be split into qualitative \cite{Portz2010} and quantitative measures \cite{Kim2012, Asano2009}. The former including assessments made by experts in the field or context of the intended application \cite{Klugl2009}, as well as category based rating systems \cite{Duives2013} designed to define the capabilities of an algorithm (such as emergent behaviours). These assess whether the simulation \emph{looks} natural and that the agents within the simulation are not acting in an unusual fashion.

A number of quantitative measures have been suggested to provide a numeric measure of accuracy for a simulation, which include but are not limited to: speed, pedestrian density, number of steps taken to destination and duration. These evaluation techniques tend to be data driven, and as such require some kind of ground truth data from which to test against. The concept of an evaluation framework has been suggested before \cite{Charalambous2014, Wolinski2014, Guy2012, Kapadia2011, Rodreiguez2011, Musse2012}; with most deducing various metrics based on a simulation in an effort to rate simulation algorithms or tune parameters. Often these frameworks evaluate the quality of a simulation based on data driven methods i.e. how closely does simulated agent A's track match pedestrian A's track in the source data. This relies on the assumption that a good simulation must mimic captured source data exactly, however, humans moving through the same environments on a regular basis will look similar but have slightly different properties, rendering this assumption flawed.

Many of these evaluation frameworks have merit in their given context, however to make a comparison, often a number of requirements are imposed on their source data. Most commonly this pertains to tracks for the pedestrians in source data, this introduces issues to the data collection process pertaining to cost, time, ethics and suitability for large outdoor environments. Additionally the focus of these frameworks is a statistical analysis of the simulations, specifically on individual agents rather that of the simulation as a whole meaning the way the simulation appears is often overlooked.

\begin{figure}
    \centering
    \includegraphics[width=0.45\columnwidth]{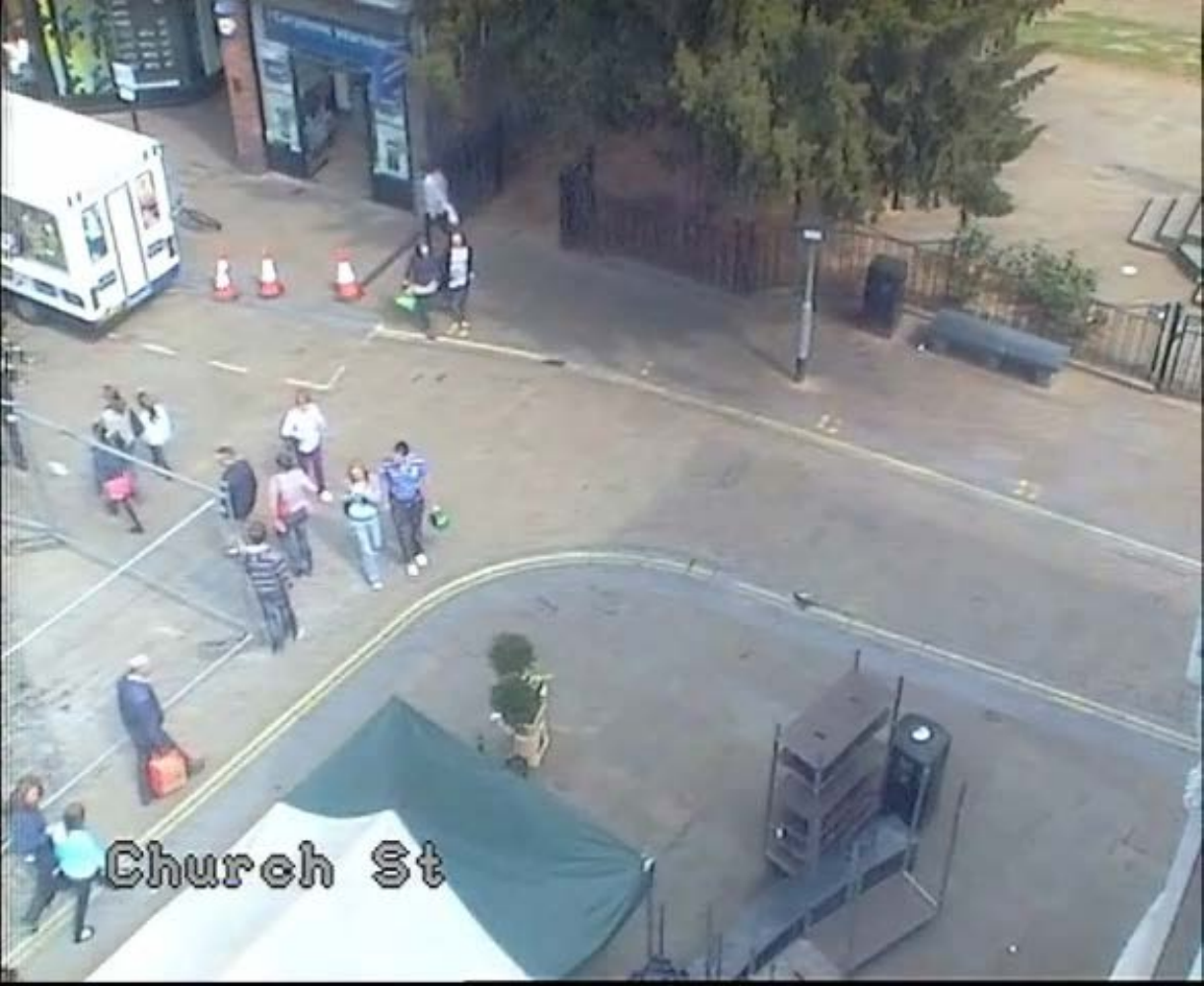}
    \includegraphics[width=0.45\columnwidth]{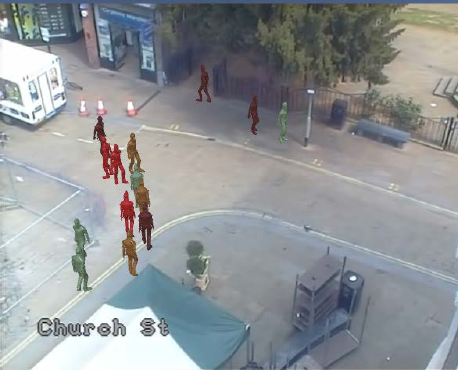}
    \caption{Source CCTV footage and generated composition video with computer controlled agents.}
    \label{fig:Initial_Comparison}
\end{figure}

To address these gaps in the existing research a new modular evaluation framework is proposed allowing comparison of a simulation algorithm to source video footage using limited ground truth data. Using novel compositing techniques, a video comprised of the 2D background of a source video and superimposed 3D agents (controlled by the simulation algorithm), can be created (Figure \ref{fig:Initial_Comparison}). Using this simulated video, a direct comparison against the source video can be performed using Human Visual System (HVS) features \cite{Jablonski2014}, analysing crowd properties such as density, speed and track. Due to the modular nature of the framework, both the composition techniques as well as the video analysis methods are interchangeable as required. This ensures the method is future proofed to new and improved methodologies.

Due to the modular nature of the framework, any simulation algorithm can be used, providing the facility to compare the performance of different simulation algorithms relative to source material. The framework also provides the functionality for model tuning, by creating a fast feedback loop which allows the adjustment of model parameters to improve simulation accuracy. Furthermore as the simulated composite videos produced as part of the proposed framework have a known groundtruth, they are very suitable for the evaluation of pedestrian tracking algorithms.

This framework and the associated features are motivated by the work in \cite{Jablonski2014} and looks to improve on the ideas proposed. As such, the following novel framework is suggested which reduces the complexity of crowd and pedestrian simulation evaluation, providing a quantitative comparison regardless of context. Simulated video sequences are created using sample video data and crowd simulation algorithms, combined using novel compositing and visualisation techniques. Through the use of background subtraction and scene composition to generate simulated video sequences, lengthy scene reconstruction steps are eliminated simplifying the overall process. The paper will continue as follows: firstly, an overview of the research in this field is given, next, the proposed modular framework and compositing methodology are introduced and an overview of HVS features presented. Finally the evaluation process is reviewed and conclusions drawn.

\section*{Literature Review}

Within this section a brief overview of current simulation algorithms is given, followed by a review of existing evaluation metrics and frameworks.
\subsection*{Simulation Algorithms}

Simulating behaviour virtually has been an area of intense research in recent years, one of the first agent-based simulation algorithms was proposed by Reynolds in \cite{reynolds1987flocks, reynolds1999} which focused on birds flocking, and was later developed and better defined for gaming applications. Later Helbing et al \cite{Helbing1998} introduces the Social Force Model (SFM). This became widely successful due to its ability to emulate common attributes seen in pedestrian movements and emergent behaviours in crowds such as line forming in tight areas. Another key advantage of this model was the use of variables that related to physical principles in our world. The use of these parameters allowed the application of other forms of research to drive the simulation and formed a basis for evaluation. Another example of this type of approach is proposed by Xi et al \cite{Xi2011}. Here a dense model is proposed integrating extended decision field theory, the social force model and a dynamic planning algorithm involving AND/OR graphs. Extensive testing is done on potential profit for a shopping mall where factors of an agents AI, for example group dynamics, visual field or intention to buy, are changed. However no real validation based on real results is presented. Survey and observation data is used in the setting of these model parameters.

With the advent of so much research in the area of pedestrian simulation, the ability to determine what is a good simulation has been well researched and is very much a topic for debate. As such work has been done to survey the current research with a number of different simulation algorithms and models being put through the same test environments \cite{Duives2013, Papadimitriou2009, singh2009, Zhan2008, Moussa2016, Gang2017}. Evaluations are based on properties such as computational performance as well as the presence of emergent behaviours and a model's abilities, such as route choice and agent strategy. Additionally numerical metrics are introduced spanning areas such as speed and acceleration, distance covered, time and collisions. However there is no real analysis of the quality of simulated visual experience, i.e if the methodology produces visually similar or natural behaviour, which is often a key requirement of the design.

In more individual cases it is seen that often the evaluation technique is limited by the context. This of course makes sense but often means that other key aspects of a simulation implementation are not analysed. For example in evacuation simulations, the ground truth data often does not exist from which to compare, leaving only qualitative assessments by experts in the field \cite{Klugl2009} leaving any other properties that these algorithms have un-explored.

Portez et al \cite{Portz2010} focuses on the simulation of crowds around bottle necks and looks for events. Here density matching against recorded video footage is used as a quantitative measurement, more specifically number of people per square meter. This is backed up using visual checks against the original footage to ensure the simulated crowds resemble those in the captured data. Pettre et al \cite{Pettre2009} introduces an agent interaction method and uses density plots based on aspects of the simulation, such as reaction times, as well as a defined likelihood function to perform their evaluation. The likelihood function is based on simulation variables and the assessed difference from captured source data, however these are only implemented on interaction scenarios comprising of two people and as such their validity as metrics for larger scale crowd simulation remains untested.

Kim et al \cite{Kim2012} manages a quantitative assessment of modelled scenarios by looking at psychological studies of specific situations. I.e looking for a simulation to repeat observations made. For example, a study of pedestrians average crossing speeds relative to the start of the crossing signal \cite{Crompton1979}. This is then compared with the outputs of the simulation to provide a similarity measure. This is a more abstract approach as they are not comparing specific paths but rather duration, focusing on the properties of the simulation rather than specific paths.

\subsection*{Simulation Evaluation Frameworks}

Charalambous et al \cite{Charalambous2014} looks at the creation of an analysis tool that characterises outlying behaviour in simulations. Two processes are suggested; outlier detection which searches for odd behaviour within that simulation and novelty detection, which finds trends or actions that differ from some reference data. However for a good analysis, reference data must be very similar to the simulated. Additionally as the comparison made is purely a data driven approach the concept of something looking similar is not addressed.

Guy et al \cite{Guy2012} uses a computed entropy score to compare simulated data to captured real world data. The metric is defined as the entropy of the distribution of errors between the evolution of a crowd predicted by a simulator and the source data. Using three differing datasets ranging from simplistic to dense crowds, evaluations are done to produce simulations that closely resemble the source data. This statistical analysis again relies on the need for position information from both the simulation and comparable real world examples. Additionally a number of assumptions are made, the most noticeable being that the crowd simulator is not systemically more accurate for some agents within a crowd than for others, however this is not always accurate as their will always be aspects of a simulation that are more accurate than others \cite{Charalambous2014}.

Wang et al \cite{Wang2016} present the Stochastic Variational Dual Hierarchical Dirichlet Process (SV-DHDP) model in which groups of similar trajectories (trending paths) can be combined to generate an overall path pattern for an environment. The path patterns created are therefore the result of local dynamics and global factors allowing differing insights based on the simulation environment. The resultant visualisations allow for detailed qualitative analysis and the introduction of an inference based similarity metric allows for the comparison of extracted path patterns from differing data sources. Providing a good generalised view of a given scene. However analysis is done on defined paths for source and test data which requires complex post processing techniques or data captured in a specific format which can create inaccuracies \cite{Asano2009}, more accurate systems can be used \cite{Wang2012a} but are not always cost effective due to the requirement of specialized equipment.

Lerner et al \cite{Lerner2009, Lerner2010} address the concept of look and feel of a crowd by assigning a similarity metric by comparing an agent's actions at a given moment in time to a database of observed actions. The database is taken from annotated frames of video, defining path vectors for pedestrians in both sparse and dense crowds. A state-action pair for each frame is defined using firstly a state (set of recorded variables such as trajectory, speed and position) and an action (a density measure representing local density changes over time). The similarity between a state-action pair from the database and a test state-action pair is defined as the similarity between the actions (differences in positions along the trajectories) and the distance between the states (differences in densities for the surrounding regions).

Musse et al \cite{Musse2012} also address the issue of tracking generalised paths in crowds using four dimensional histograms to describe movement within a crowd. By applying the Bhattacharyya distance as a form of measurement between the defined features similarity is assessed on criteria such as speed, spacial occupancy and orientation. The results produce similarity measurements for aspects of orientation and speed but fail to take into account the density of the crowds. Additionally no analysis of what is visually similar is given which highlights a systemic problem with many of the similarity features suggested such that evaluation is given based on similarity to extracted trajectories but not on if the results look the same to a group of people.

Additionally the work in \cite{Rodreiguez2011, Banerjee2011} proposes interesting similarity metrics and \cite{Kapadia2011, Zanlungo2014} are also worth mentioning.

The proposed framework is focusing on providing a statistical analysis of the realism in different simulations. In order to evaluate the realism of a crowd or pedestrian simulation algorithm, vision based features are utilised. Motion estimation \cite{Argyriou2007,Argyriou2011} and tracking are a few of the vision based steps applied during the process of pedestrian and crowd behavior analysis for visual surveillance in dynamic scenes \cite{Bloom2015,Bloom2014}.

The majority of today's optical flow methods strongly resemble the original formulation provided by Horn and Schunck \cite{Horn1981} as well as the work by Lucas and Kanade \cite{Lucas1981}. The accuracy and robustness of optical flow estimation algorithms has seen significant improvement over the last decade \cite{Sun2010, Sun2014, Makris2002}. Additionally work in tracking pedestrians and crowds has also seen much work \cite{Munder2008, Raptis2010}. Specifically techniques have been developed for estimating the flux of people in public areas, such as stores or travel sites, which can then automatically provide congestion analysis assisting in management of crowds and pedestrians \cite{Allain2009, hu2004Survey, lakoba2005}.

A technique that incorporates optical flow for accuracy evaluation in crowd simulation was proposed in \cite{clarke2007}. In this work a solution is proposed which allows the relationship of optical flow to physical velocity to be defined. The main issue of this approach is that it requires manual annotation and performs well only for specific relative orientations of the camera and pedestrians.


All these approaches for pedestrian and crowd simulation evaluation either do not consider the quality of simulated visual experience, or fail to compare their given metric with human validation. Additionally the requirement for data with a defined groundtruth introduces issues pertaining to cost, time, ethics and suitability for large outdoor environments. All key components when considering the accessibility and accuracy of crowd simulation evaluation.

\section*{Methodology}

\subsection*{Simulation Evaluation using Compositing Techniques} \label{sec:5_Sim_Evaluation}

The following section will explain in detail the various aspects of the proposed modular Crowd Simulation Evaluation through Composition (CSEC) framework. The framework provides a method of simulation algorithm evaluation which rates how realistic simulated human walking behaviours look compared to sample footage. Evaluation can be done on a frame by frame basis or on a sequence as a whole, providing flexibility in how the simulation is evaluated. Additionally the proposed methodology requires no track or path information for the source material, allowing any pedestrian video footage captured from a static viewpoint to be used as source material. Evaluation of an algorithm's performance is key to defining how realistic the simulation outputs are.


Comparison is made using the original source footage and a video created using composition techniques. This utilises background subtraction techniques as well as methods to extract 3D data from 2D images. This allows the construction of a 3D space in which virtual agents can navigate around. Through the use of composition, a final visualisation combining this background and 3D space is generated to form the simulated video sequence in which simulated agents are superimposed into the background of the source video data. Analysis of these generated videos is through the use of features designed to evaluate the visual similarity of the two videos to provide a quantifiable similarity metric. These are designed to emulate the way the Human Visual System (HVS) perceives motion, and include the principles of Weber's Law \cite{Weber1834} to better match the metrics to the way humans see.

\begin{figure*}
    \centering
    \includegraphics[width=15cm]{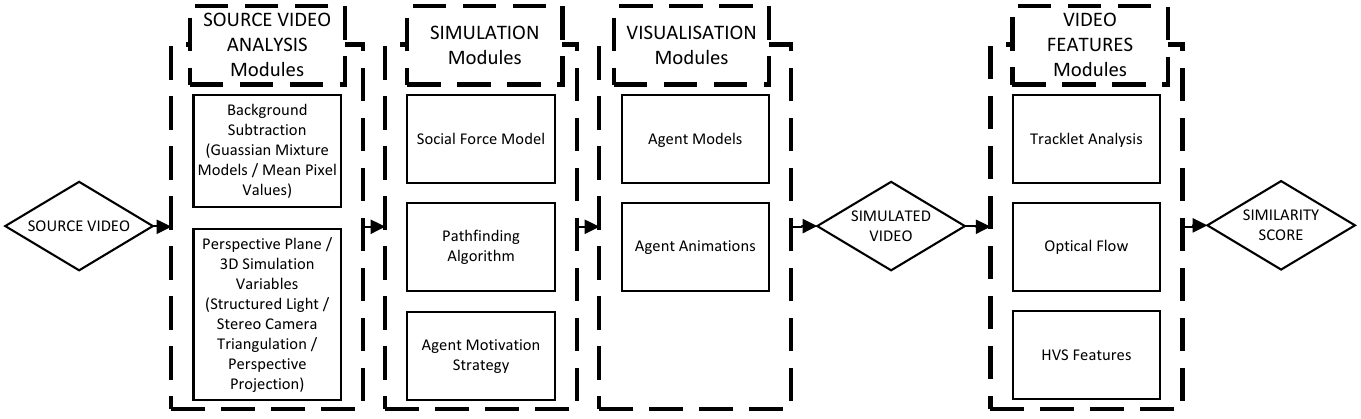}
    \caption{Outline of the modular nature of the Crowd Simulation Evaluation through Composition (CSEC) framework. Using an initial source video, an analysis stage produces the inputs for simulation and visualisation. Simulation generates the paths that the agents will follow and the visualisation modules combine these aspects to create the simulated video. The video features module is concerned with the comparative aspect of the framework and the similarity score created as an output.}
    \label{fig:5_Modular_Framework}
\end{figure*}

\begin{figure}
    \centering
    \includegraphics[width=0.95\columnwidth]{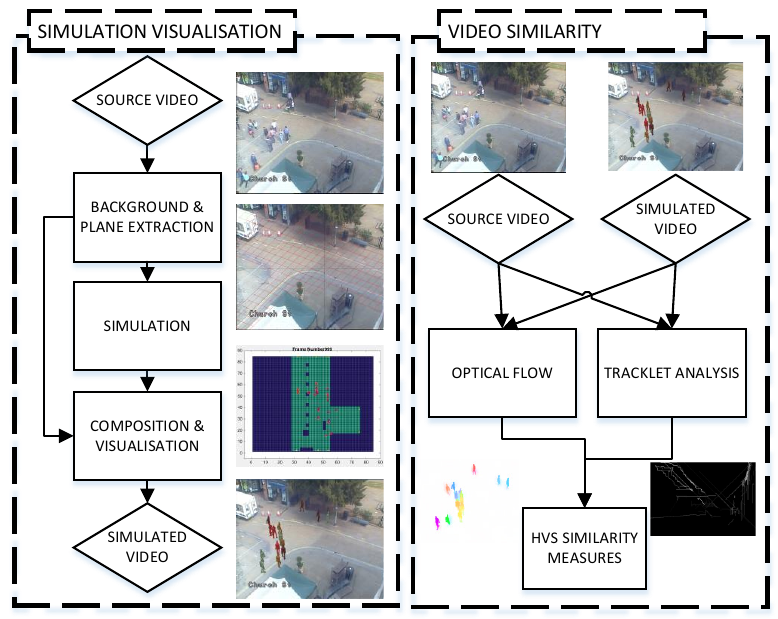}
    \caption{Overview of the Crowd Simulation Evaluation through Composition (CSEC) framework.}
    \label{fig:5_Overview}
\end{figure}

Fundamentally the framework is made up of two components; simulation visualisation and video similarity. Importantly the modular nature of the framework supporting inputs of any simulation algorithm or video analysis techniques, depending on application, whilst still retaining the ability to produce a quantifiable similarity score (Figure \ref{fig:5_Modular_Framework}). The HVS features \cite{Jablonski2014} provide a good generalisation of simulation evaluation requirements in a broad range of situations, however additional feature descriptors could be developed and inserted into the framework to determine other crowd statistics such as lane formation and direction of travel. Figure \ref{fig:5_Overview} provides a more specific overview of the CSEC framework as it is utilised in this work. Further detail of each section will follow.

\begin{figure}
    \centering
    \includegraphics[width=0.45\columnwidth]{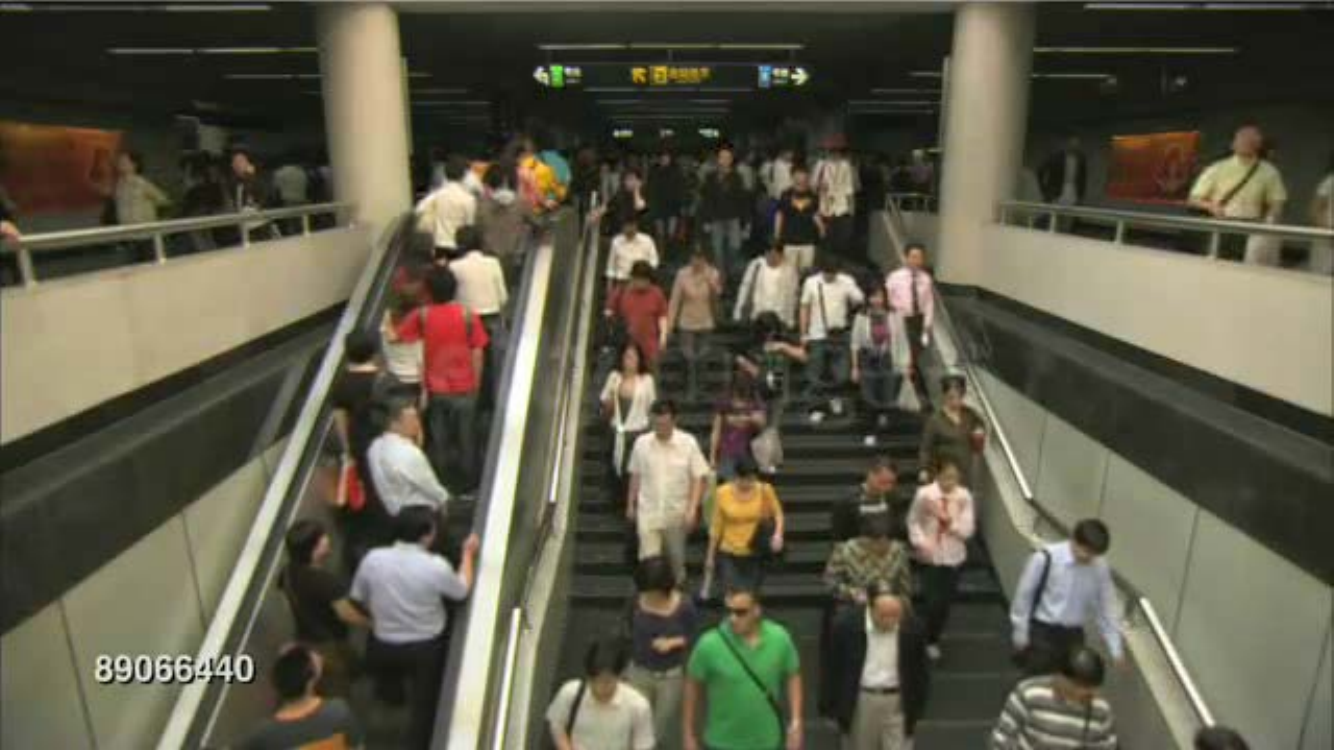}
    \includegraphics[width=0.45\columnwidth]{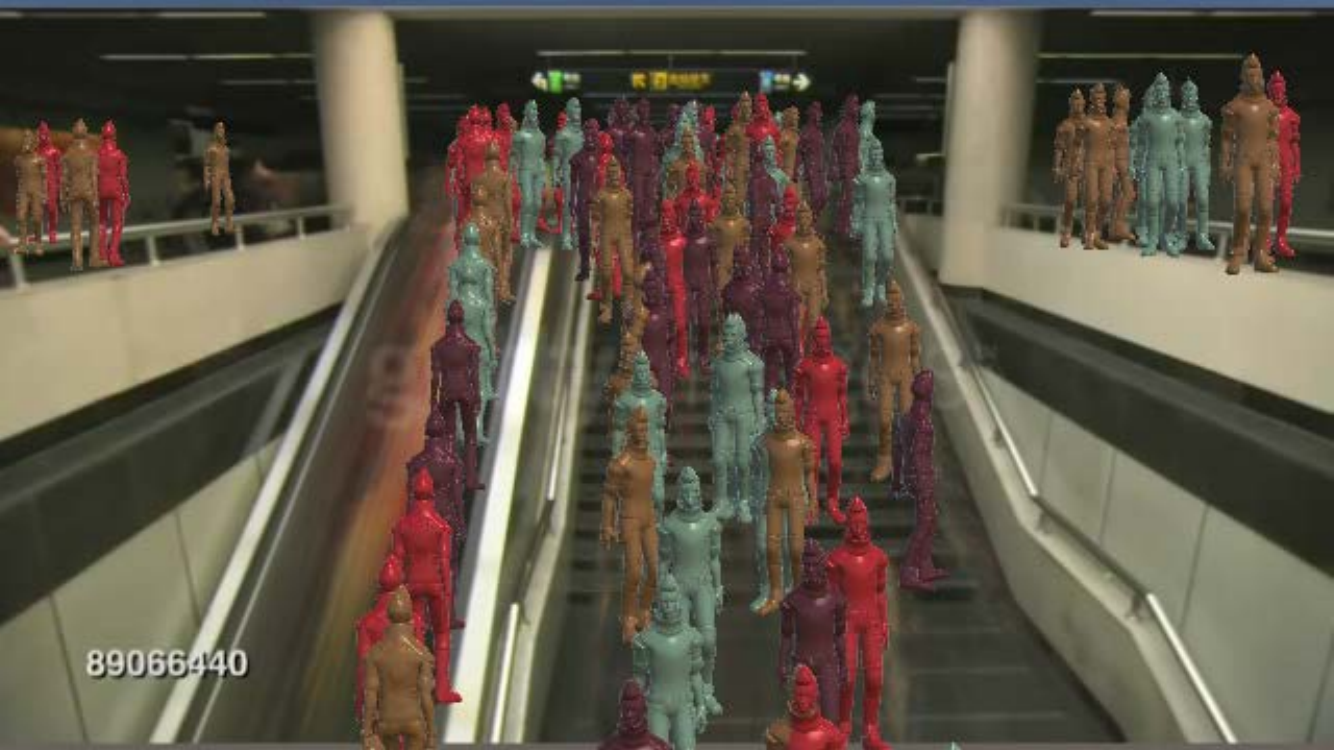}
    \caption{Frames of source CCTV footage and generated video using the composition techniques.}
    \label{fig:5_Initial_Comparison}
\end{figure}

As the framework compares video data to derive a similarity value, firstly a simulated video must be constructed. Initially, using the source video sequence, the background is extracted. Next, a two dimensional plane is extracted representing a top down view of the given scene. Simulations are run to produce paths for virtual agents to follow based on the extracted plane. The visualisation component is used to composite the extracted 2D background image and 3D rendered agents as they follow the simulated paths. Frames are output from the visualisation into a final simulated video sequence (Figure \ref{fig:5_Initial_Comparison}). Once both a simulated and source video are available, the similarity can be evaluated. Optical flow and tracklet analysis are run and features extracted from the subsequent data. Finally a distance measure is used to evaluate the difference in features to give the final similarity metric.

\subsubsection*{Background and Plane Extraction} \label{sec:5_Background_Plane_Extract}

To allow the composition of the simulated video to be created, the background of the source video sequence is required. For this work the mean value of each pixel in a video sequence is used to create a compound background image. Other methods based on Gaussian mixture models could also be used in order to obtain more accurate results \cite{Zivkovic2004}.

\begin{figure}[!ht]
    \centering
    \includegraphics[width=0.45\columnwidth]{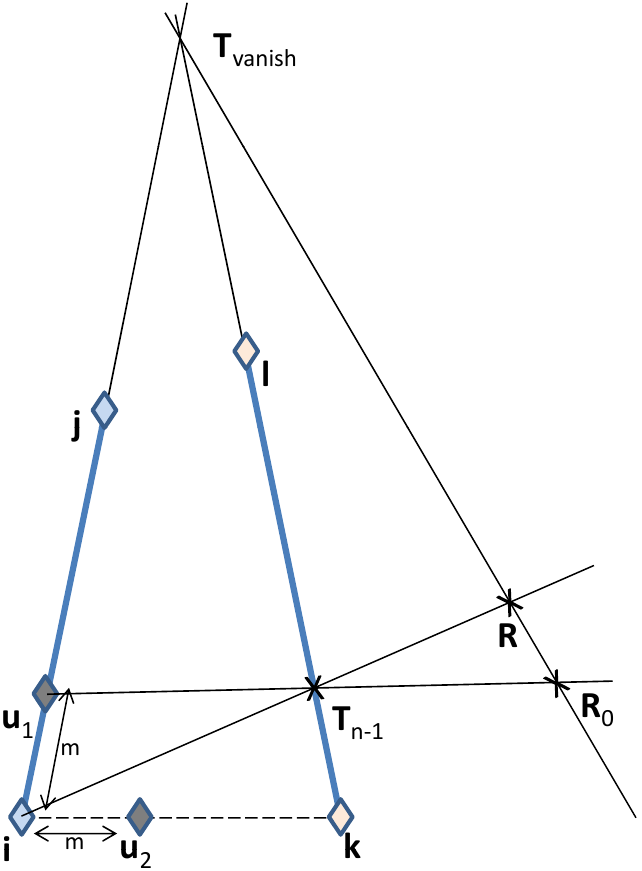}
    \includegraphics[width=0.45\columnwidth]{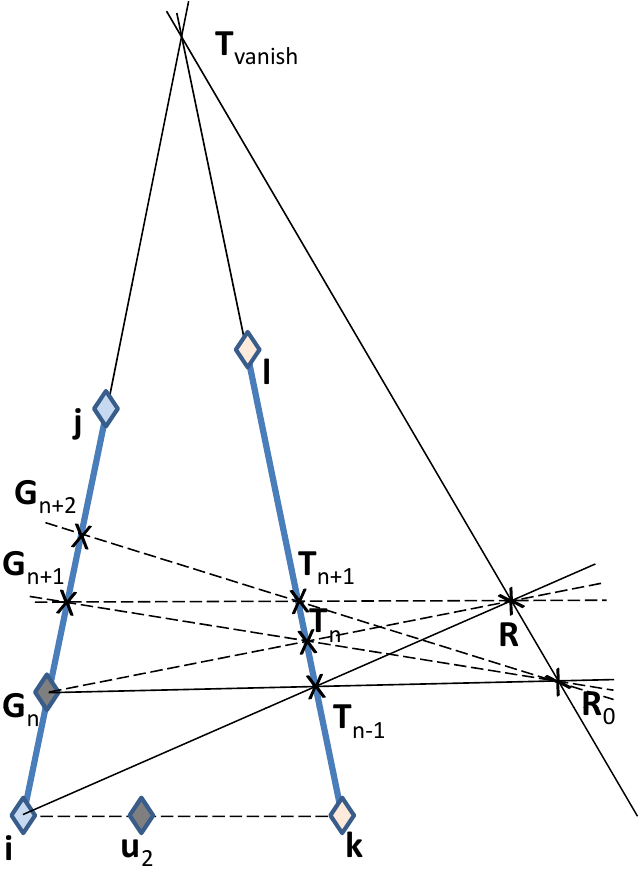}
    \caption{(a) User defined points and initialisation. (b) The first two iterations of the recursive algorithm. }
    \label{fig:PerspectiveGrid}
\end{figure}

Once the background image has been subtracted the process of defining the perspective grid is applied. The perspective grid allows scale mapping of an environment from the viewpoint of the source video camera pose. The resultant grid represents a top down environment map of the viewable area and is used during agent simulation. Using the concept of perspective scale along a line we can, through the definition of two parallel lines that run to the vanishing point of an image, estimate distance in arbitrary units of measure within this perspective space (Figure \ref{fig:PerspectiveGrid}b). This unit can be based upon an object in the scene with known dimensions or using pedestrians \cite{Chan2008}.

Initially the user defines the points $\mathbf{i}$ and $\mathbf{j}$, in the 2D image space, forming a line along an edge that leads to the vanishing point of the image. A second line is defined by the points $\mathbf{k}$ and $\mathbf{l}$, such that it runs `parallel', relative to the vanishing point in the 3D space of the captured image, to the line defined by points $\mathbf{i}$ and $\mathbf{j}$ (Figure \ref{fig:PerspectiveGrid}a).

At a location along the line $\mathbf{i} \mathbf{j}$ the user defines another point $\mathbf{u}_1$, such that the line $\mathbf{i} \mathbf{u}_1$ represents the unit of distance $m$ from which all further perspective points are defined. An additional point $\mathbf{u}_2$ is defined on top of the line $\mathbf{i} \mathbf{k}$ which represents the same relative distance in 3D space as $m$.

For the next step of the proposed algorithm the reference points $\mathbf{T}_{vanish}$, $\mathbf{R}$, $\mathbf{R}_0$ and $\mathbf{T}_{n-1}$ are initialised automatically (Figure \ref{fig:PerspectiveGrid}a).

In more detail, the vanishing point $\mathbf{T}_{vanish}$ is defined as the point at which the lines $\mathbf{i} \mathbf{j}$ and $\mathbf{k} \mathbf{l}$ intersect, this may well be at a position outside of the image plane. As such $\mathbf{T}_{vanish}$ is defined as

\begin{align}
    \mathbf{T}_{vanish} = f(\mathbf{i}, \mathbf{j},  \mathbf{k}, \mathbf{l})
\end{align}

An arbitrary point $\mathbf{R}$ is selected at a random location outside the triangle $\mathbf{i} \mathbf{T}_{vanish} \mathbf{k}$. The point $\mathbf{T}_{n-1}$ is defined as the point of intersection of the lines $\mathbf{i} \mathbf{R}$ and $\mathbf{k} \mathbf{T}_{vanish}$

\begin{align}
    \mathbf{T}_{n-1} = f(\mathbf{i}, \mathbf{R},  \mathbf{k}, \mathbf{T}_{vanish})
\end{align}

Finally the point $\mathbf{R}_0$ is defined.

\begin{align}
    \mathbf{R}_0 = f(\mathbf{u}_1, \mathbf{T}_{n-1}, \mathbf{R}, \mathbf{T}_{vanish})
\end{align}


With these points initialised a recursive algorithm is applied to calculate equidistant points along the line $\mathbf{i} \mathbf{T}_{vanish}$ in 3D space. As the user has already defined the first of these points $\mathbf{u}_1$, for the purposes of the recursive step, these will be relabeled as $\mathbf{G}_n$.
This is a two-step iterative process, with the point $\mathbf{T}_n$ being defined as the intersection of the lines $\mathbf{G}_n \mathbf{R}$ and $\mathbf{k} \mathbf{T}_{vanish}$.

\begin{align}
    \mathbf{T}_n = f(\mathbf{G}_n, \mathbf{R}, \mathbf{k},\mathbf{T}_{vanish})
\end{align}
and during the second step the next equidistant point $\mathbf{G}_{n+1}$  on the line $\mathbf{i} \mathbf{T}_{vanish}$ is expressed as a function of

\begin{align}
    \mathbf{G}_{n+1} = f(\mathbf{R}_0, \mathbf{T}_n,  \mathbf{i}, \mathbf{T}_{vanish})
\end{align}

This process is repeated until $\mathbf{G}_{n+1}$ is no longer within the borders of the original background image.

The grid is initially defined using all the equidistant points on the line $\mathbf{i} \mathbf{k}$ using the distance $\mathbf{i} \mathbf{u}_2$ as a unit. Lines are defined between each of these points and the vanishing point $\mathbf{T}_{vanish}$ of the image. The scale points $\mathbf{G}$ are plotted along each of these newly defined lines forming the grid. Additionally if required, the recursive process can be inverted to create points moving away from the vanishing point. This ensures that the entire image plane is encapsulated by the defined grid, regardless of where the user has defined their points.

\begin{figure}
    \centering
    \includegraphics[width=0.9\columnwidth]{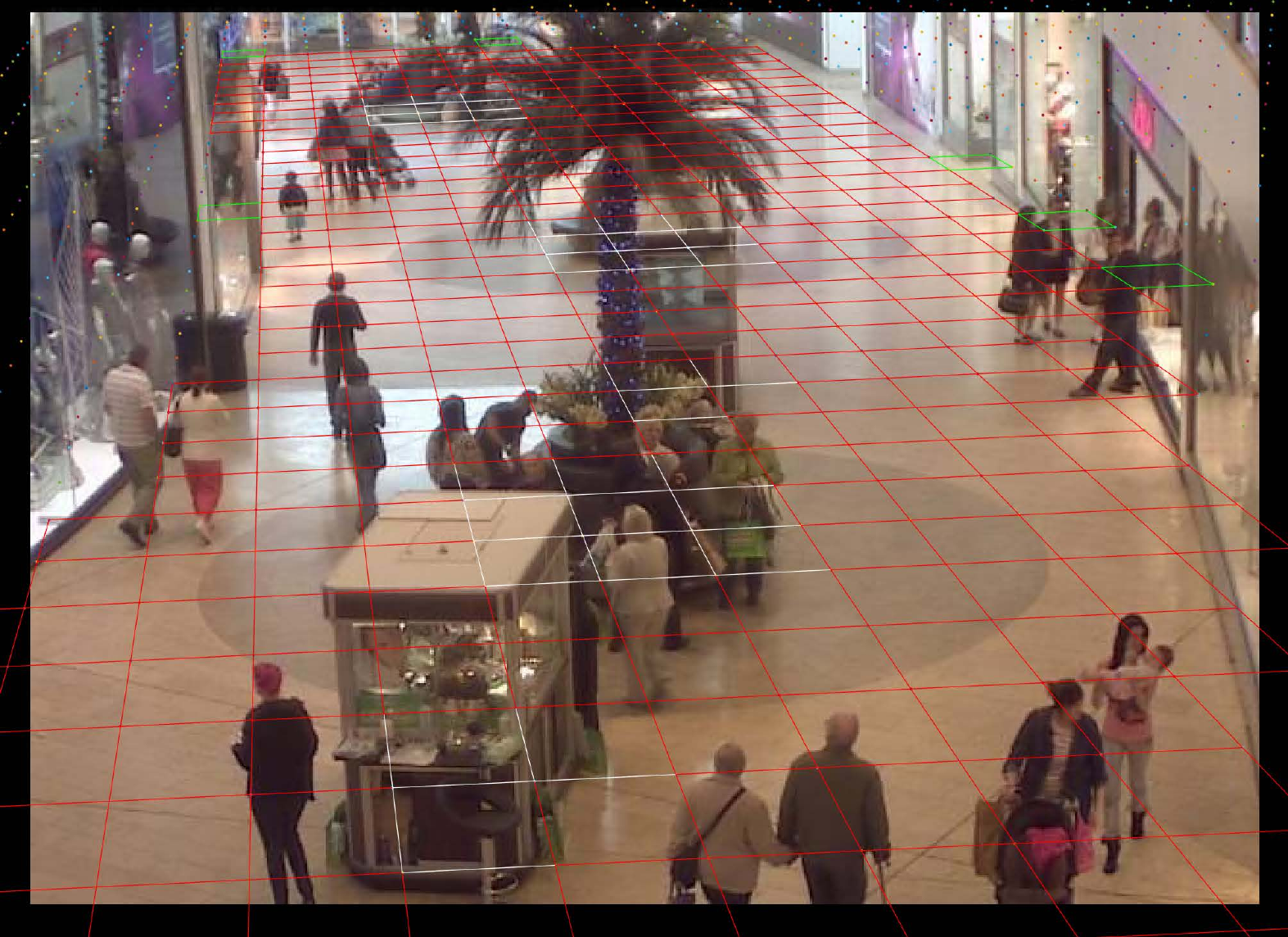}
    \caption{Resultant perspective grid overlayed on the original source image. Red cells indicate areas where agents can walk, white represent obstacles and green marks an entrance or exit.}
    \label{fig:5_Perspective_Grid}
\end{figure}

The resultant grid represents the perspective plane of the source image. On that grid the areas (cells) with obstacles (i.e. cells where pedestrians cannot walk) are annotated as is information about entrance/exit locations. In order to help the user; the obtained grid is superimposed on the extracted background image (Figure \ref{fig:5_Perspective_Grid}). Here red cells indicate areas where agents can walk, white represent obstacles and green marks an entrance or exit. This annotated version of the perspective plane is then used as the ground plane by the simulation algorithms.

\subsection*{Pedestrian and Crowd Simulation Algorithm} \label{sec:5_Simulation}

To simulate the agent movement through a given scene an algorithm based on a combination of simulation methodologies is used. Firstly a steering simulator based on the work of \cite{reynolds1999} in which the concepts of simple crowd behaviours such as separation, object avoidance and agent collision detection are utilised. These have been implemented with the social forces model structure in which each of these elements produce a force applied to the agent to adjust their movement vector. The magnitude of these forces is based on distance. An additional step, using a planning simulation methodology, based on the work of Karamouzas et al \cite{Karamouzas2009} is used as a predictive collision detection algorithm to produce natural agent avoidance within the simulations, this again is implemented by the application of a force upon the simulated agent. As outlined in (\ref{equ:5_Forces}).

\begin{align}
F_a = g_a - p_a + \sum^{n}_{i = 1}f(a,b_{i}) + \sum^{m}_{j = 1}f(o_{j}) + \sum^{o}_{k = 1}f(a,b_{k})
\label{equ:5_Forces}
\end{align}
where $g_a$ is the current destination along the path of the agent $a$ to its final goal, with $p_a$ being the agent's current position. The forces for separation, $f(a,b_{i})$, object avoidance $f(o_{j})$ and the predicative agent avoidance $f(a,b_{k})$, is calculated for any relevant entity within a defined neighbourhood. For overall path planning, an agent performs a route plan using the A* algorithm and the perspective plane obtained previously to estimate the most direct course from their start location to their destination. The variables associated with defining an agent and their respective movements are based on existing work used by Asano et al. \cite{Asano2009,Asano2010} who in turn derive their values from the existing studies and from observational data from their datasets.

\subsection*{Composition and Visualisation}

The visualisation stage of the framework performs the composition of a scene utilising the extracted background obtained from the source video and the generated perspective plane. The key to a visually similar composition is the positioning of a virtual camera at the same location as in the original scene. By using layers the camera can have the source image as a background and the visualised 3D agents controlled by the simulation superimposed. Due to this, it is important to line up the perspective plane with the background to give the illusion of the agents walking through the scene. This alignment can be performed manually using the position and orientation of the camera or automatically using camera calibration techniques \cite{Guo2014, Zeisl2015}.

\begin{figure}
  \centering
  \includegraphics[width=0.9\columnwidth]{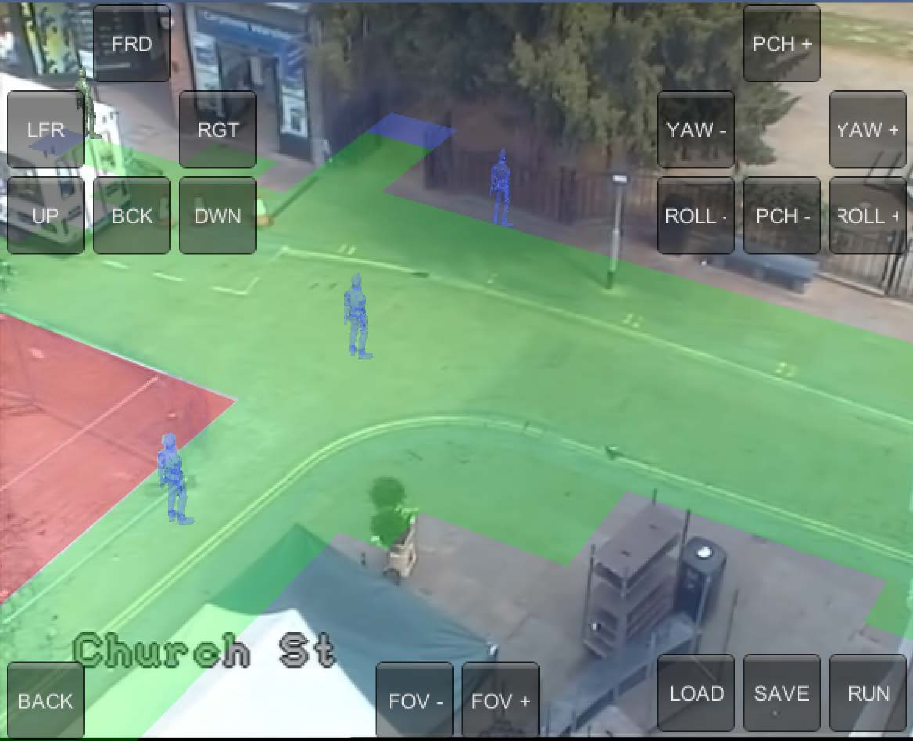}
\caption{Example composition of the Kvan scene with test agents and perspective floor plan.}
\label{fig:5_Visual}
\end{figure}
Sample agents are then placed in the scene at various locations to ensure that the perspective and scaling parameters of the agents are appropriate to the scene. Figure \ref{fig:5_Visual} demonstrates this with agents in blue positioned at different locations in the scene. These values can be adjusted manually or calculated automatically using the methods provided in \cite{Do2013}. In Figure \ref{fig:5_Visual} it can also be seen that the imported floor plan is coloured according to each cell's defined values, green meaning areas the agents can walk and blue defines entrance exit points and red represents obstructions that will obscure the agents when they are located behind. To control the agents in the scene, position and orientation information is required for each frame. This is obtained using the desired simulation algorithm and the same perspective grid map outlined in the Simulation section.

As the goal is to create videos with similar crowds, a number of parameters from the original video are required. Using the source video sequence, an analysis is made of the pedestrians in the scene, outlining paths and estimated crowd density. For this work the information was extracted manually or provided by the datasets used, however work exists to help automate this process \cite{Wang2016, Rodreiguez2011}. Once all the required parameters for the agents are defined the simulation is run and outputs recorded at the frame rate of the original source video.

Finally, with the composition completed and the simulations run the visualisation of the scene is performed. Agent models are created and sized according to the obtained parameters. For each frame of the simulation the agent location and rotation is updated based on the simulation algorithm output and a composite frame is captured. Once visualisation is completed the individual frames are compiled into a video sequence. Importantly the resolution and the number of frames in the new composite video should be equal to that of the source video.

\subsection*{Simulation Similarity Metrics}
Once visualisation of the composite video is complete, the source and the simulated video sequences are used to extract features in order to measure their level of similarity. These features are based on the optical flow and tracklets of the moving objects in both sequences.

An optical flow method tries to calculate the motion between two image frames at times $t$ and $t + \delta t$ at each pixel position \cite{Lucas1981}. The solution as given by Lucas and Kanade is a non-iterative method, which assumes a locally constant flow in a small window. Black and Anandan in \cite{Sun2010}, describe how the single motion assumption, as well as the constant brightness constraint are not always valid. They discuss how these assumptions can be relaxed in order to develop a more robust estimation framework.

Tracklet estimation is a well researched topic with many algorithms available in the literature. These can be based on motion or other features and utilise particle and Kalman filters \cite{Munder2008, Raptis2010, KhoongW2011,Hu2004}. Specifically, the problem of motion based tracking can be split into detecting moving objects in each frame and the association of those moving elements to a continuous corresponding object over time.

In the case of Kalman filters, the track's location in each frame is predicted and a likelihood of a detection is assigned to each track. The Kalman filter is a recursive estimator, meaning that only the estimated state from the previous time step and the current measurement are needed for computation of the current state. The Kalman filter has two distinctive features; firstly its mathematical model is described in terms of state-space concepts; Secondly, the solution is computed recursively. Usually the Kalman filter is described by a system state model and a measurement model.

In this system the optical flow algorithm proposed in \cite{Sun2010} and the tracking method presented in \cite{KhoongW2011} were utilized, however the system is designed in such a way that allows the incorporation of multiple motion estimation or tracking methods as plugins. Based on this system architecture the proposed evaluation framework is dynamic and capable of utilizing current and future state of the art tracking methods.

\subsubsection*{Motion and Tracklet Flux Similarity Metrics} \label{sec:5_Metrics}

In order to evaluate the similarity level of the simulated and source videos a new metric is required that will allow an objective comparison incorporating the Human Visual System (HVS) based similarity features. Weber's Law \cite{Weber1834} and the work in \cite{Wharton2008, zanker1995} states that a human's ability to define motion, as the point when the signal-to-noise ratio is regarded as at a stimulus intensity. Therefore, the minimum motion contrast $dV$ as a function of background motion $V$, required for the human visual system to notice a change is expressed as:

\begin{align}
    dm = L \frac{dV}{V}
    \label{equ:5_Weber}
\end{align}

where $dm$ is the differential change in motion perception, $dV$ is the differential increase in the velocity and $V$ is the velocity. The parameter $L$ is to be estimated using experimental data. The proposed measure includes Fechner’s Law, which relates velocity $V$, to perceived motion, $\mathbf{M}$, as seen by the human visual system, as follows:

\begin{align}
    \mathbf{M} = Lln(\frac{V}{V_{max}})
\end{align}

where $V_{max}$ is the ‘upper threshold’ of the human eye. The proposed metric is based on the motion and tracklet flux histograms obtained from the perceived motion $\mathbf{M}$ utilizing standard computer vision algorithms.

Let us assume that $I_R(\vec{u},t)$ and $I_S(\vec{u},t)$ are the image frames of a real and the correspondent simulated scene, respectively. The motion vectors for each pixel location in each frame are estimated using the optical flow techniques shown in (\ref{equ:5_OF1}) and (\ref{equ:5_OF2}).

\begin{align}
    M_R (\vec{u},t)=f(I_R (\vec{u},t),I_R (\vec{u},t-1)) \label{equ:5_OF1} \\
    M_S (\vec{u},t)=f(I_S (\vec{u},t),I_S (\vec{u},t-1)) \label{equ:5_OF2}
\end{align}

The estimated tracklets are obtained using motion information and Kalman filters.

\begin{align}
    T_R (n_R,\vec{u},t)=f(M_R, I_R) \label{equ:5_TRACK1} \\
    T_S (n_S,\vec{u},t)=f(M_S, I_S) \label{equ:5_TRACK2}
\end{align}

Since the motion vectors and the tracklets are available the Histogram of Oriented Optical Flow (HOOF) \cite{Chaudhry2009} is calculated both for the real and simulated scenes.

\begin{align}
    f_R^{HOOF}=HOOF(M_R) \label{equ:5_HOOF1} \\
    f_S^{HOOF}=HOOF(M_S) \label{equ:5_HOOF2}
\end{align}

Also, a 2D histogram of the motion parameters is obtained using (\ref{equ:5_2DH1}) and (\ref{equ:5_2DH2}).

\begin{align}
    & f_R^{H2D} (r_{ij})=m_{ij} (M_R) \label{equ:5_2DH1}\\
    & f_S^{H2D} (r_{ij})=m_{ij} (M_S) \label{equ:5_2DH2}
\end{align}
where $r_{ij}$ is the $i^{th}$ and $j^{th}$ motion level in an interval, and $m_{ij}$ is the number of pixels in all the given frames whose motion level is $r_{ij}$. Regarding the tracklets, the time parameter in (\ref{equ:5_TRACK1}) and (\ref{equ:5_TRACK2}) is removed by superimposing all of them at the same time instance. The similarity metric here can be applied on any given time interval, which can be the whole sequence or a small time fragment. In the same way as in (\ref{equ:5_2DH1}) and (\ref{equ:5_2DH2}) we obtain:

\begin{align}
    f_R^T (r_{ij})=m_{ij} (T_R) \\
    f_S^T (r_{ij})=m_{ij} (T_S) \label{equ:5_TRACK4}
\end{align}

Finally, the flux of the features in (\ref{equ:5_HOOF1}) - (\ref{equ:5_TRACK4}) is represented by the surface integral of the given vector field.

\begin{align}
    \Phi(\vec{u},t)=\Sigma_{\vec{u}} \Sigma_t f dudt
    \label{equ:5_FLUX}
\end{align}

Based on (\ref{equ:5_FLUX}), we obtain $\Phi_R^{HOOF}$, $\Phi_S^{HOOF}$, $\Phi_R^{H2D}$, $\Phi_S^{H2D}$, $\Phi_R^T$ and $\Phi_S^T$ that correspond to the proposed HVS features. All the features can be applied either on the whole sequence or on smaller blocks allowing specio-temporal adaptation of the proposed features and metrics. In order to measure the similarity and rank the algorithms, the Bhattacharyya distance is utilised due to its use in similar work \cite{Musse2012}.

\section*{Results}

To evaluate the proposed Crowd Simulation Evaluation through Composition (CSEC) framework, a total of five different scenes were used from various crowd datasets (Mall Dataset \cite{Loy2013}, PETS2009 \cite{PETS2009} and RBK \cite{Simonnet2011}) and captured crowd and pedestrian videos sequences. Scenes of different environments including both indoor and outdoor spaces, with a large range of camera orientations and crowd configurations. Additionally the frame rates of the videos varied from less than $10$fps up to $24$fps providing a challenging and diverse set of scenes from which to evaluate the effectiveness of the evaluation framework.
\begin{figure}
    \centering
    \includegraphics[width=0.45\columnwidth]{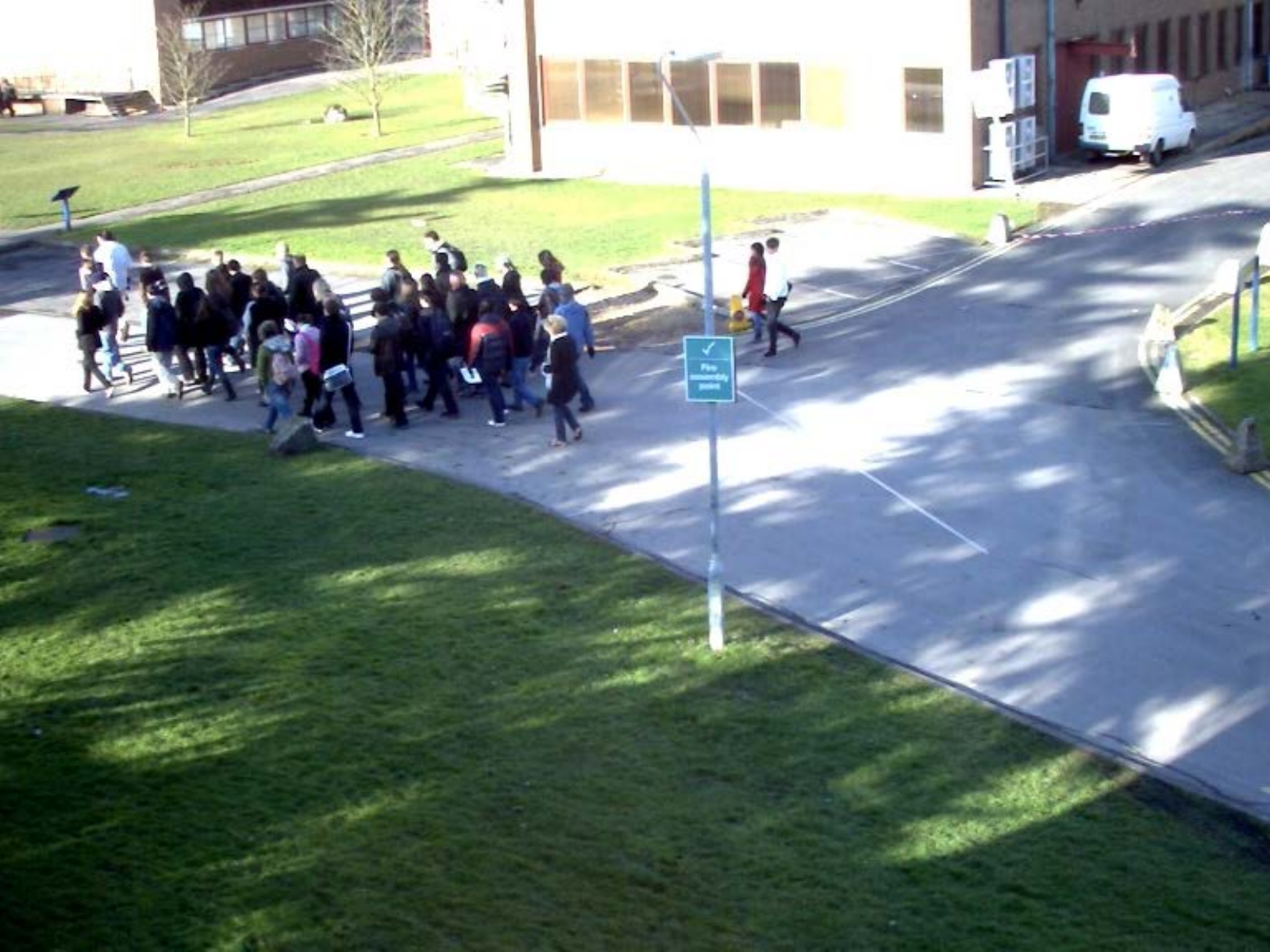}
    \includegraphics[width=0.45\columnwidth]{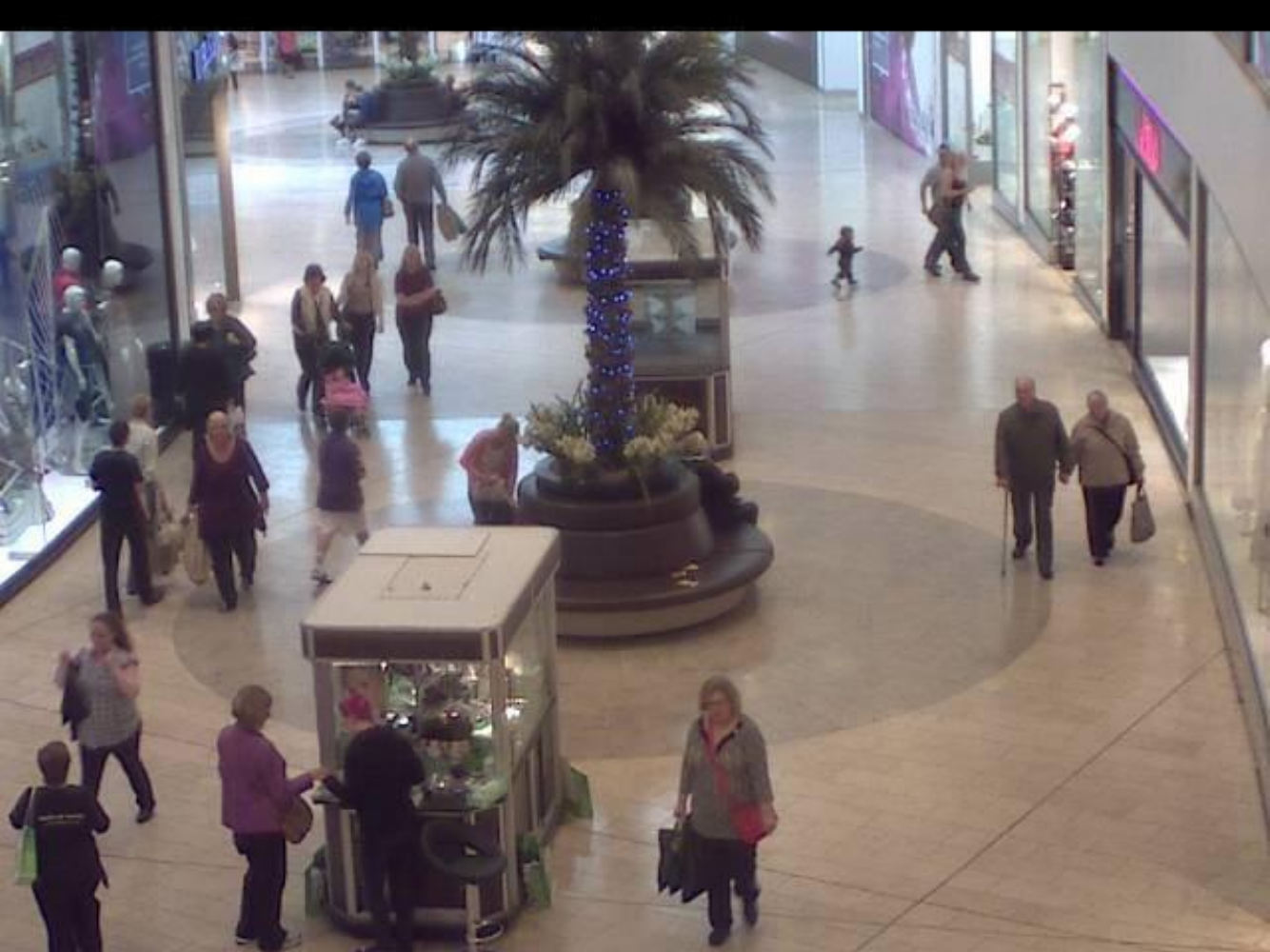}\\
    \includegraphics[width=0.45\columnwidth]{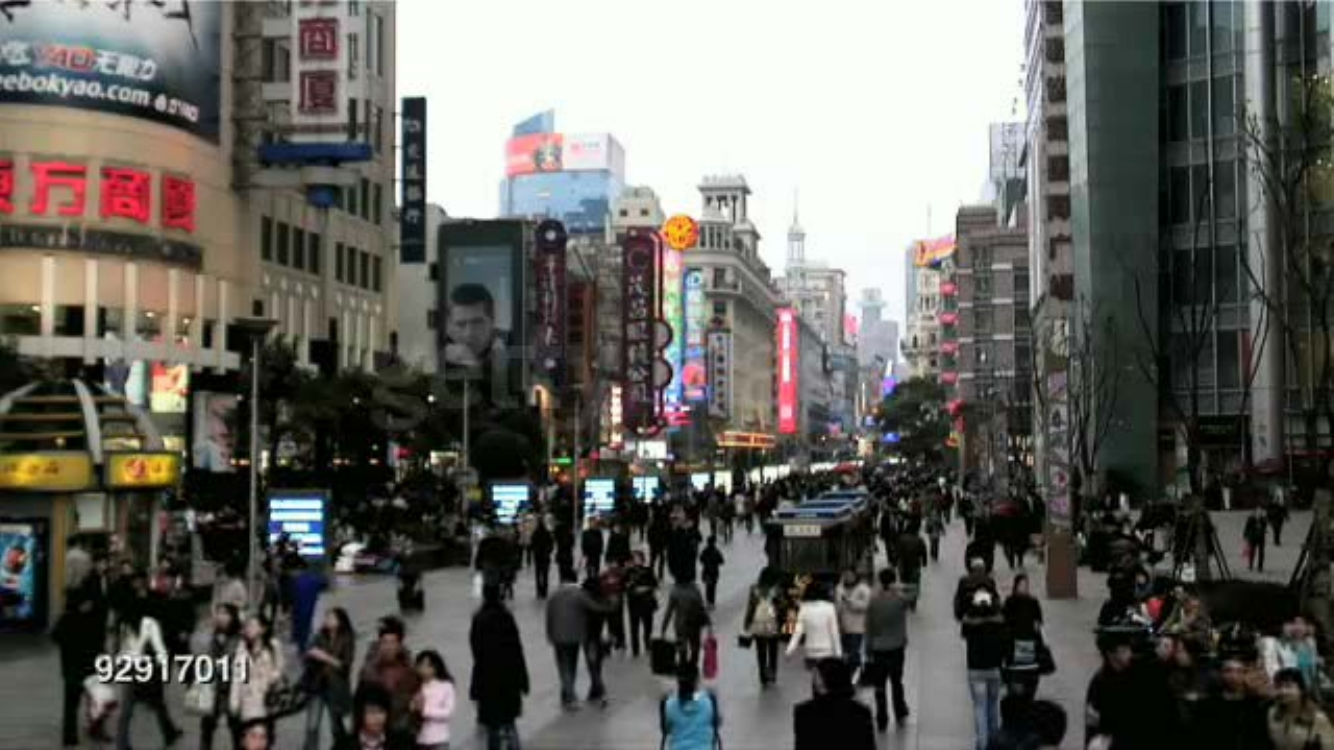}
    \includegraphics[width=0.45\columnwidth]{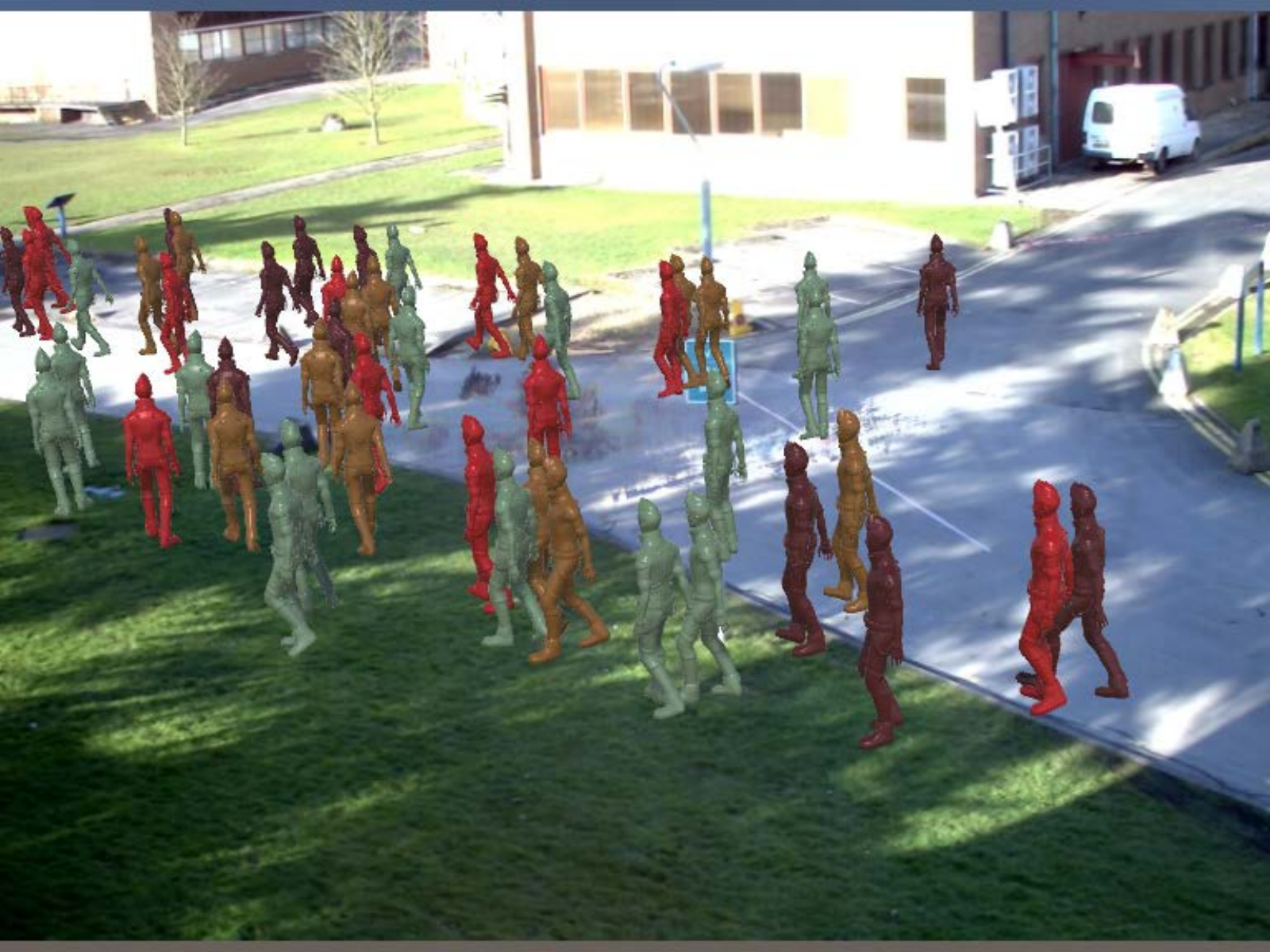}\\
    \includegraphics[width=0.45\columnwidth]{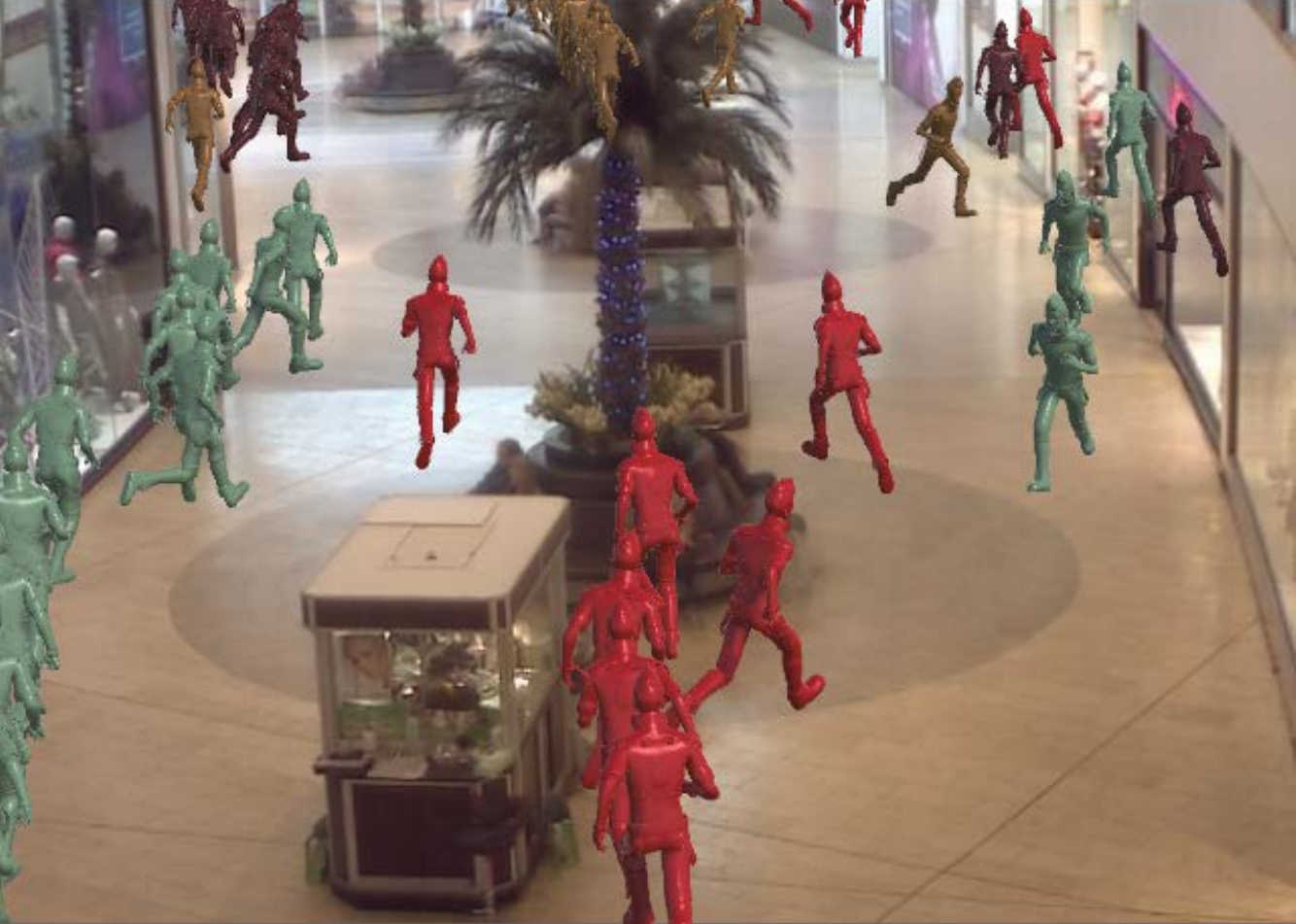}
    \includegraphics[width=0.45\columnwidth]{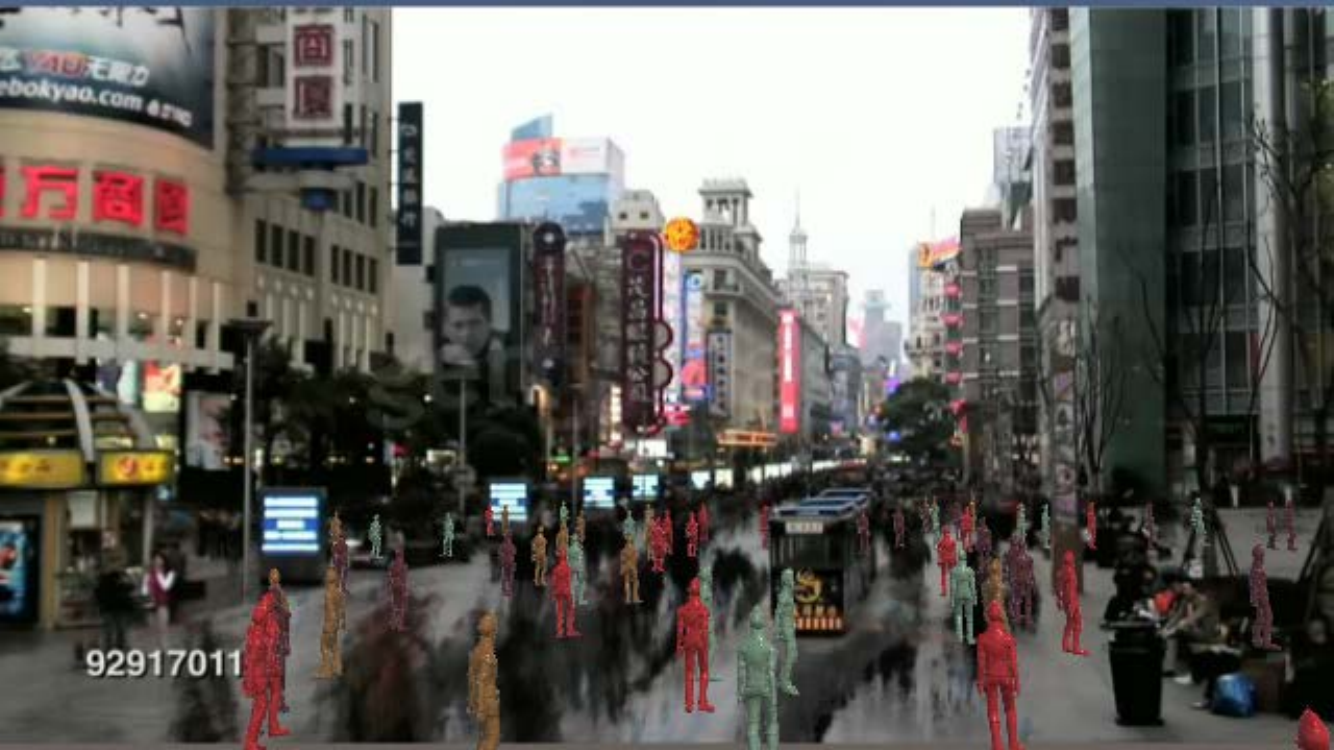}
    \caption{Example source and simulated frames. Top Row (Source): Road, Mall, Krad2. Bottom Row (Simulated): Road, Mall, Krad2.}
    \label{fig:Scenes}
\end{figure}

Composite simulated videos for each of the tested scenes were created using four different levels of agent speed and three different population levels, totalling 12 simulations and therefore 12 composite videos per scene. This demonstrates the framework's ability to evaluate source footage against a single simulation algorithm. An additional 12 composite videos were created using Reynolds' flocking algorithm \cite{reynolds1987flocks}, providing a comparative test between simulation algorithms and demonstrating a comprehensive assessment of the relative features of the framework. Figure \ref{fig:Scenes} presents example source and simulated frames for a few scenes.

For each scene the background image was extracted and the perspective grid defined. Simulations were performed for each of the cases mentioned previously and the outputs used to create composite visualisations for each. Simulated videos were then created with the same frame rate, length and resolution as the source videos. The simulations themselves are run at a set frame rate ($50$ frames per second), a desired frame rate is also specified allowing for the movement of the agents to be visualised at the same frame rate as the source video without having to adjust the agent properties between simulations.

Each simulated video, and its respective source video, had the optical flow and tracklets estimated. Finally the HVS features \cite{Jablonski2014} were used to compare each visualisation against its source. Three features were used in the comparison, the tracklets ($\Phi_R^T$ (\ref{equ:5_TRACK1}) and $\Phi_S^T$ (\ref{equ:5_TRACK2})), Histogram of Orientated Optical Flow per frame ($\Phi_R^{HOOF}$ (\ref{equ:5_HOOF1}), $\Phi_S^{HOOF}$ (\ref{equ:5_HOOF2})) and the Histogram of Orientated Optical Flow for the sequence ($\Phi_R^{H2D}$ (\ref{equ:5_2DH1}), $\Phi_S^{H2D}$ (\ref{equ:5_2DH2})).

The HOOF features and 2D Histograms used a window size of $64 \times 64$ pixels per frame. Using these features, a generalised statistical measure of the differences in movement from the source human behaviour to the simulated agents is defined. The distance metric used to compare the features is the Bhattacharyya Distance \cite{bhattachayya1943} due to its use in similar work. For these experiments no pre-defined groundtruth is required, instead each scene has the number of agents and their speed estimated. It is expected that simulations that have a similar number of agents and relative speed to the source video will produce the lowest distance measure.

Tables \ref{tab:5_Average_HOOF} - \ref{tab:5_Average_Distance} (left side) contain the average distance measures, after applying equation (\ref{equ:5_FLUX}), across all tested scenes for the 12 composite videos using the algorithm outlined in the Pedestrian and Crowd Simulation Algorithm section. As expected the lowest distance values are seen when the simulation parameters closely match those of the source material. Also in Table \ref{tab:5_Average_Distance} the average feature combination results are given for the composite videos generated using the Reynolds flocking algorithm \cite{reynolds1987flocks}, in these simulations the same initialisation values and constraints were applied as in the previous results. The cells of the table are coloured green-yellow-red, whereby green is a low distance and therefore a close match to the source footage and red represents a large measured distance and therefore less accurate simulation. Therefore Table \ref{tab:5_Average_Distance} demonstrates a direct comparison where it can be seen that the distance measures from the source footage for the Reynolds \cite{reynolds1987flocks} algorithm are consistently higher than seen in the comparative results from the proposed simulation algorithm. This is logical as modern alternatives to Reynolds \cite{reynolds1987flocks} flocking algorithm should present more realistic movement. This demonstrates that the suggested framework provides a useful comparison tool from which to analyse simulation similarity.

\begin{table}
\centering
\caption{Average Bhattacharyya distance between source and each simulated video sequence using the simulation algorithm outlined in the Simulation section, across all five Scenes for $\Phi^{HOOF}$ features.}
\label{tab:5_Average_HOOF}
\begin{tabular}{c|cccl|}
\cline{2-5}
                                & \multicolumn{4}{c|}{Agents Speed}                                                               \\ \hline
\multicolumn{1}{|c|}{\# Agents} & Very Slow                  & Slow                        & \textbf{Same}               & Fast   \\ \hline
\multicolumn{1}{|c|}{Few}       & \multicolumn{1}{c|}{9.32} & \multicolumn{1}{c|}{9.07} & \multicolumn{1}{c|}{8.08} & 8.01 \\ \cline{2-5}
\multicolumn{1}{|c|}{\textbf{Same}}& \multicolumn{1}{c|}{9.42} & \multicolumn{1}{c|}{8.41} & \multicolumn{1}{c|}{\underline{7.33}} & 7.91 \\ \cline{2-5}
\multicolumn{1}{|c|}{Many}      & \multicolumn{1}{c|}{9.41} & \multicolumn{1}{c|}{8.83} & \multicolumn{1}{c|}{8.95} & 9.47  \\ \hline
\end{tabular}
\end{table}

\begin{table}
\centering
\caption{Average Bhattacharyya distance between source and each simulated video sequence using the simulation algorithm outlined in the Simulation section, across all five Scenes for $\Phi^{H2D}$ features.}
\label{tab:5_Average_H2D}
\begin{tabular}{c|cccl|}
\cline{2-5}
                                & \multicolumn{4}{c|}{Agents Speed}                                                            \\ \hline
\multicolumn{1}{|c|}{\# Agents} & Very Slow                  & Slow                       & \textbf{Same}              & Fast  \\ \hline
\multicolumn{1}{|c|}{Few}       & \multicolumn{1}{c|}{2.50} & \multicolumn{1}{c|}{2.19} & \multicolumn{1}{c|}{\underline{2.10}} & 2.24 \\ \cline{2-5}
\multicolumn{1}{|c|}{\textbf{Same}}& \multicolumn{1}{c|}{2.63} & \multicolumn{1}{c|}{2.52} & \multicolumn{1}{c|}{2.44} & 2.55 \\ \cline{2-5}
\multicolumn{1}{|c|}{Many}      & \multicolumn{1}{c|}{2.86} & \multicolumn{1}{c|}{2.69} & \multicolumn{1}{c|}{2.71} & 2.88 \\ \hline
\end{tabular}
\end{table}

\begin{table}
\centering
\caption{Average Bhattacharyya distance between source and each simulated video sequence using the simulation algorithm outlined in the Simulation section, across all five Scenes for $\Phi^T$ features.}
\label{tab:5_Average_Tracklet}
\begin{tabular}{c|cccl|}
\cline{2-5}
                                & \multicolumn{4}{c|}{Agents Speed}                                                            \\ \hline
\multicolumn{1}{|c|}{\# Agents} & Very Slow                 & Slow                      & \textbf{Same}                          & Fast  \\ \hline
\multicolumn{1}{|c|}{Few}       & \multicolumn{1}{c|}{4.90} & \multicolumn{1}{c|}{3.33} & \multicolumn{1}{c|}{2.49}     & 3.34 \\ \cline{2-5}
\multicolumn{1}{|c|}{\textbf{Same}}      & \multicolumn{1}{c|}{4.59} & \multicolumn{1}{c|}{2.91} & \multicolumn{1}{c|}{\underline{1.45}} & 3.48 \\ \cline{2-5}
\multicolumn{1}{|c|}{Many}      & \multicolumn{1}{c|}{6.19} & \multicolumn{1}{c|}{4.33} & \multicolumn{1}{c|}{4.95}     & 3.89 \\ \hline
\end{tabular}
\end{table}

\begin{table*}[t]
\centering
\caption{The average Bhattacharyya distance between source and each simulated video sequence using the simulation algorithm outlined in the Simulation section and, for comparison Reynolds \cite{reynolds1987flocks}, across all five scenes for the feature combination. Results have been colour coded to provide a heat map of similarity, where green represents low distance from source video and red high.}
\label{tab:5_Average_Distance}
\begin{tabular}{c|cccc|cccc|l}
\cline{2-9}
                                               & \multicolumn{4}{c|}{Proposed Method}                                                                                      & \multicolumn{4}{c|}{Reynolds \cite{reynolds1987flocks}}                                                                                                              \\ \cline{2-9}
\multicolumn{1}{c|}{\cellcolor[HTML]{FFFFFF}} & \multicolumn{8}{c|}{Agents Speed}                                                                                                                                                                                                                                         \\ \cline{1-9}
\multicolumn{1}{|c|}{\# Agents}               & V. Slow                    & Slow                         & \textbf{Medium}              & Fast                         & V. Slow                    & Slow                         & \textbf{Medium}                                     & Fast      &                   \\ \cline{1-9}
\multicolumn{1}{|c|}{Few}                      & \cellcolor[HTML]{FFCB2F}3.15 & \cellcolor[HTML]{15B616}2.45 & \cellcolor[HTML]{15B616}2.44 & \cellcolor[HTML]{15B616}2.47 & \cellcolor[HTML]{009901}2.82 & \cellcolor[HTML]{FFCB2F}3.38 & \cellcolor[HTML]{FFCC67}3.60                        & \cellcolor[HTML]{15B616}2.97 & \\
\multicolumn{1}{|c|}{\textbf{Same}}            & \cellcolor[HTML]{FFCB2F}3.11 & \cellcolor[HTML]{32CB00}2.63 & \cellcolor[HTML]{646809}2.22 & \cellcolor[HTML]{32CB00}2.74 & \cellcolor[HTML]{FFCC67}3.49 & \cellcolor[HTML]{FFCB2F}3.66 & \cellcolor[HTML]{F74944}{\color[HTML]{333333} 3.80} & \cellcolor[HTML]{32CB00}3.27 & \\
\multicolumn{1}{|c|}{Many}                     & \cellcolor[HTML]{CB0000}4.01 & \cellcolor[HTML]{FFCB2F}3.06 & \cellcolor[HTML]{FFCB2F}3.29 & \cellcolor[HTML]{FFCB2F}3.09 & \cellcolor[HTML]{FD6864}3.73 & \cellcolor[HTML]{FFCB2F}3.67 & \cellcolor[HTML]{FE0000}3.85                        & \cellcolor[HTML]{FFCB2F}3.66 & \multirow{-4}{*}{\includegraphics[height=1.7cm]{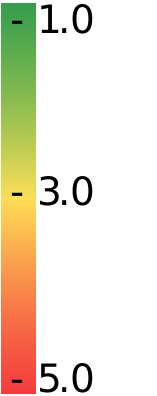}} \\ \cline{1-9}
\end{tabular}
\end{table*}

To further evaluate the methodology, a group of ten people were asked to rate the 12 simulated visualisations created using the simulation algorithm outlined in the Pedestrian and Crowd Simulation Algorithm Section, against their respective source material. Focus was given to evaluating the speed, number and track of the agents in each video compared to the source. Using the Mean Opinion Score (MOS) method, the participants were asked to provide a rating of one to five where five represented a high similarity to the source material and one a strong dissimilarity. The values for all the participants were averaged to give a score for each scene. This evaluation technique demonstrates how similarly the proposed features and metrics react compared with the participants Human Visual Systems. The results are contained in Table \ref{tab:5_MOS_Distance}, here the same green-yellow-red colour scheme is employed to allow comparison with Table \ref{tab:5_Average_Distance}, in these cases it is expected that a similar colour distribution should be seen between the tables.

\begin{table}
\centering
\caption{Mean Opinion Score (MOS) of human observations of similarity. Results have been colour coded to provide a heat map of similarity, where green represents high MOS for observed similarity to source video and red low. Provides a visual comparison between the MOS scores and the frameworks distance metrics. }
\label{tab:5_MOS_Distance}
\begin{tabular}{c|cccc|l}
\cline{2-5}
                                                        & \multicolumn{4}{c|}{Speed of Agents}                                                                                      \\ \cline{1-5}
\multicolumn{1}{|c|}{\cellcolor[HTML]{FFFFFF}\# Agents} & Very Slow                    & Slow                         & Medium                       & Fast      &                   \\ \cline{1-5}
\multicolumn{1}{|c|}{Few}                               & \cellcolor[HTML]{CB0000}1.02 & \cellcolor[HTML]{FFCC67}2.95 & \cellcolor[HTML]{32CB00}3.35 & \cellcolor[HTML]{FE996B}2.32 &  \\
\multicolumn{1}{|c|}{Same}                              & \cellcolor[HTML]{FF2A2A}1.92 & \cellcolor[HTML]{32CB00}3.27 & \cellcolor[HTML]{646809}4.36 & \cellcolor[HTML]{FFCC67}2.87 & \\
\multicolumn{1}{|c|}{Many}                              & \cellcolor[HTML]{FD6864}1.53 & \cellcolor[HTML]{32CB00}3.18 & \cellcolor[HTML]{32CB00}3.45 & \cellcolor[HTML]{FFCE93}2.72 & \multirow{-4}{*}{\includegraphics[height=1.7cm]{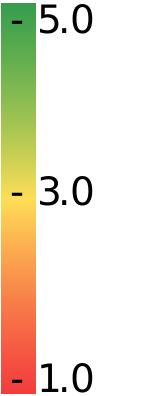}} \\ \cline{1-5}
\end{tabular}
\end{table}

The left three columns of Table \ref{tab:5_Correlation} is a breakdown of individual features against the human participant's ability to evaluate video properties. It can be seen that in certain instances the correlation between human and specific feature types is reasonably high. However by using the weighted sum of all three proposed metrics, and again comparing to the MOS, a more robust methodology is seen. This is not surprising as it is often observed that humans have difficulty distinguishing the difference between large amounts of slow moving agents versus a smaller amount of agents moving faster. As a result the combination of distance metrics from all three features more closely matches the Human Visual System's ability to evaluate motion. The weighting of the combination in this case is equal, however the optimal combination will be application dependant. Some metrics will perform better on different types of scenario. For example videos recorded from a lower point of view may not return descriptive tracklet information. To better match the HVS feature outputs with human perception Weber's Law is incorporated (\ref{equ:5_Weber}), the right three columns of Table \ref{tab:5_Correlation} demonstrate the improvement seen to the MOS correlations by doing so. In all cases the combination of metrics better correlates to the human perception of movement in the videos.

\begin{table}
\centering
\caption{Correlation (Pearson) between combination features distance and MOS, with and without Weber's Law applied.}
\label{tab:5_Correlation}
\begin{tabular}{c|c|c|c|c|c|c|}
\cline{2-7}
                                    & \multicolumn{3}{c|}{Without Weber}                & \multicolumn{3}{c|}{With Weber}       \\ \hline
\multicolumn{1}{|c|}{Metric}        & Avg               & Agts         & Spd           & Avg        & Agts         & Spd       \\ \hline
\multicolumn{1}{|c|}{$\Phi^{HOOF}$} & 0.67              & 0.49          & 0.70          & 0.54      & 0.31          & 0.62      \\
\multicolumn{1}{|c|}{$\Phi^T$}      & 0.59              & 0.46          & 0.60          & 0.63      & 0.59          & 0.57      \\
\multicolumn{1}{|c|}{$\Phi^{H2D}$}  & 0.24              & 0.06          & 0.41          & 0.28      & 0.02          & 0.44      \\ \hline
\multicolumn{1}{|c|}{Comb} & \multicolumn{1}{c|}{0.55} & \multicolumn{1}{c|}{0.36} & \multicolumn{1}{c|}{0.60} & \underline{0.61}      & \underline{0.44}          & \underline{0.65}     \\ \hline
\end{tabular}
\end{table}

\begin{figure}
    \centering
    \includegraphics[width=0.45\columnwidth]{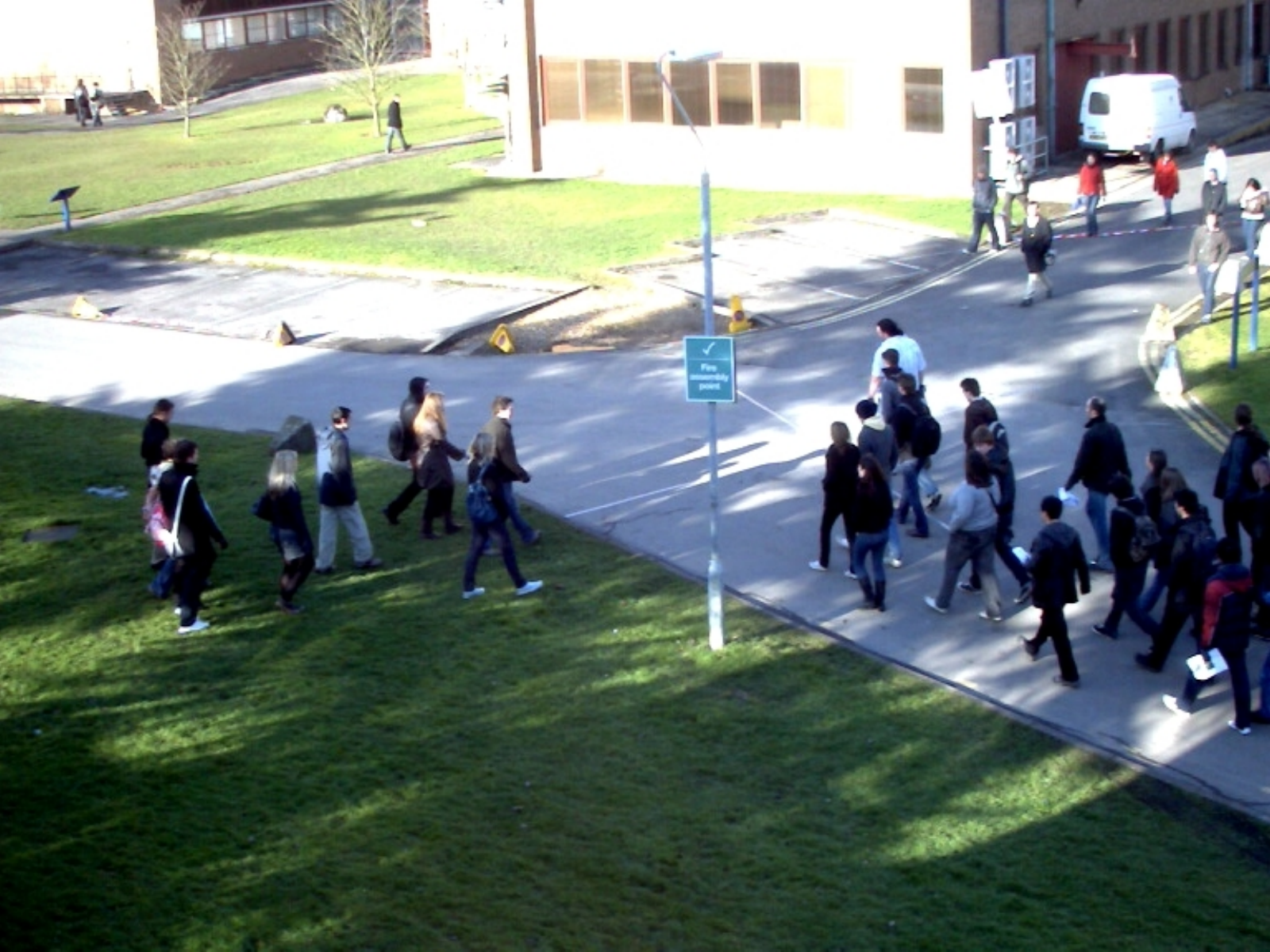}
    \includegraphics[width=0.45\columnwidth]{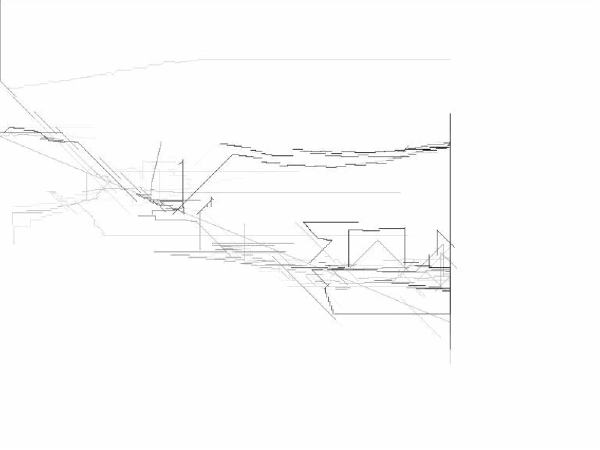}\\
    \includegraphics[width=0.45\columnwidth]{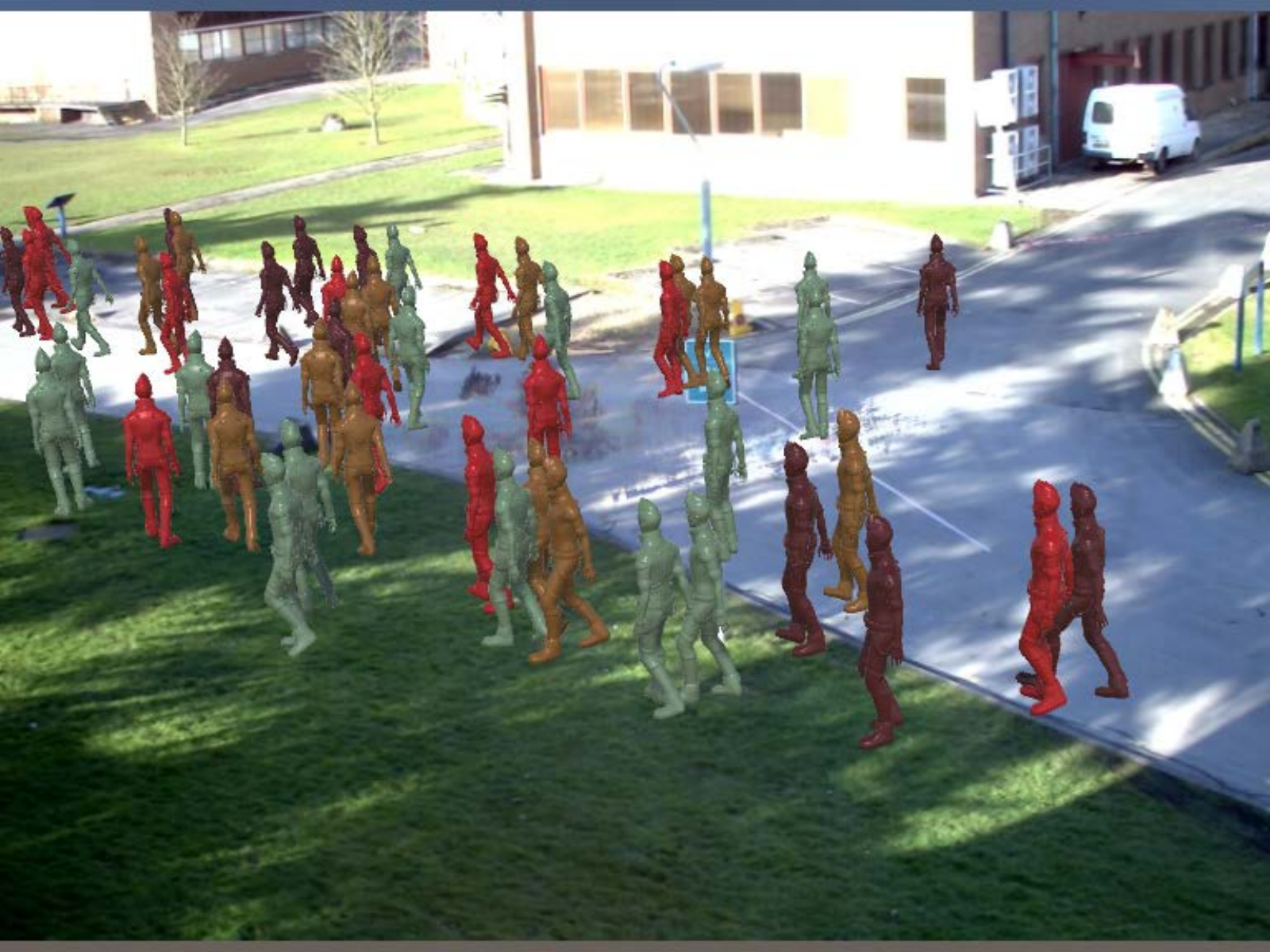}
    \includegraphics[width=0.45\columnwidth]{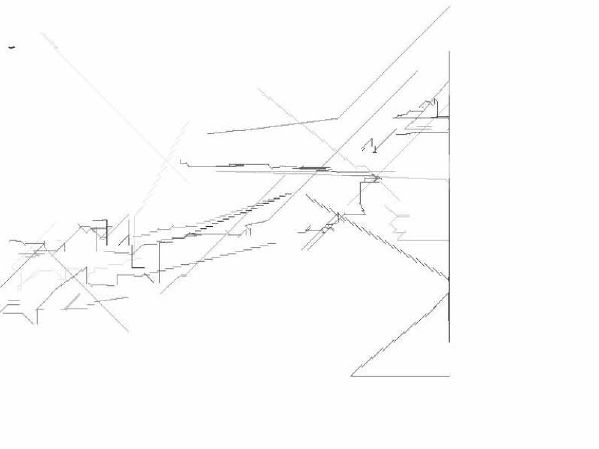}\\
    \includegraphics[width=0.45\columnwidth]{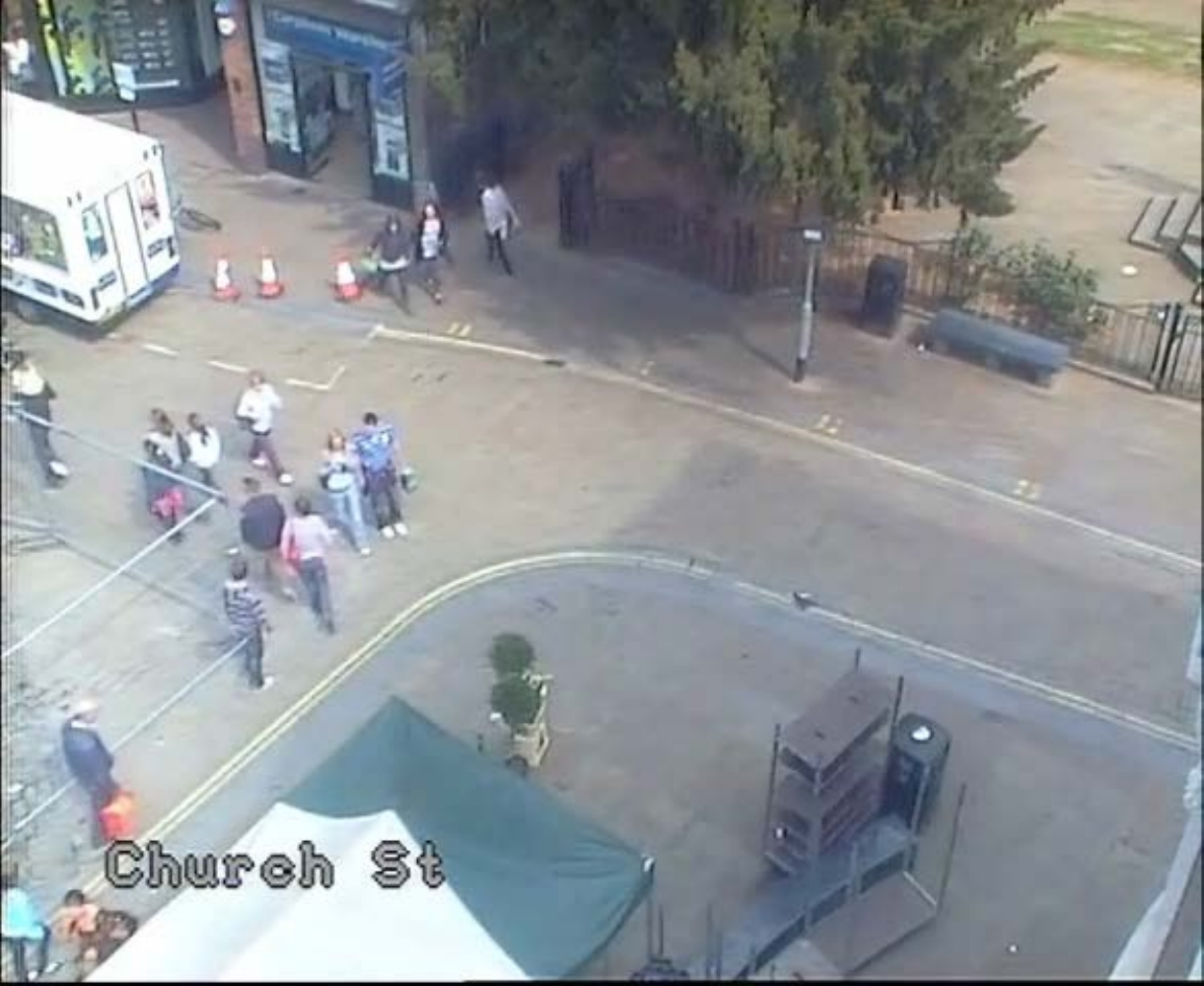}
    \includegraphics[width=0.45\columnwidth]{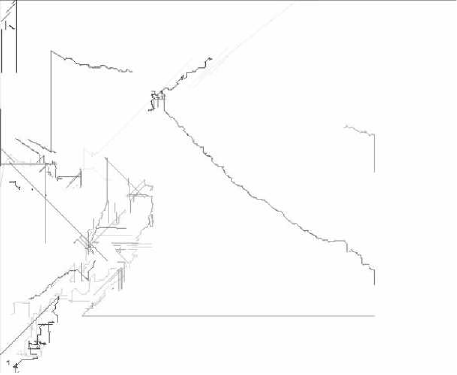}\\
    \includegraphics[width=0.45\columnwidth]{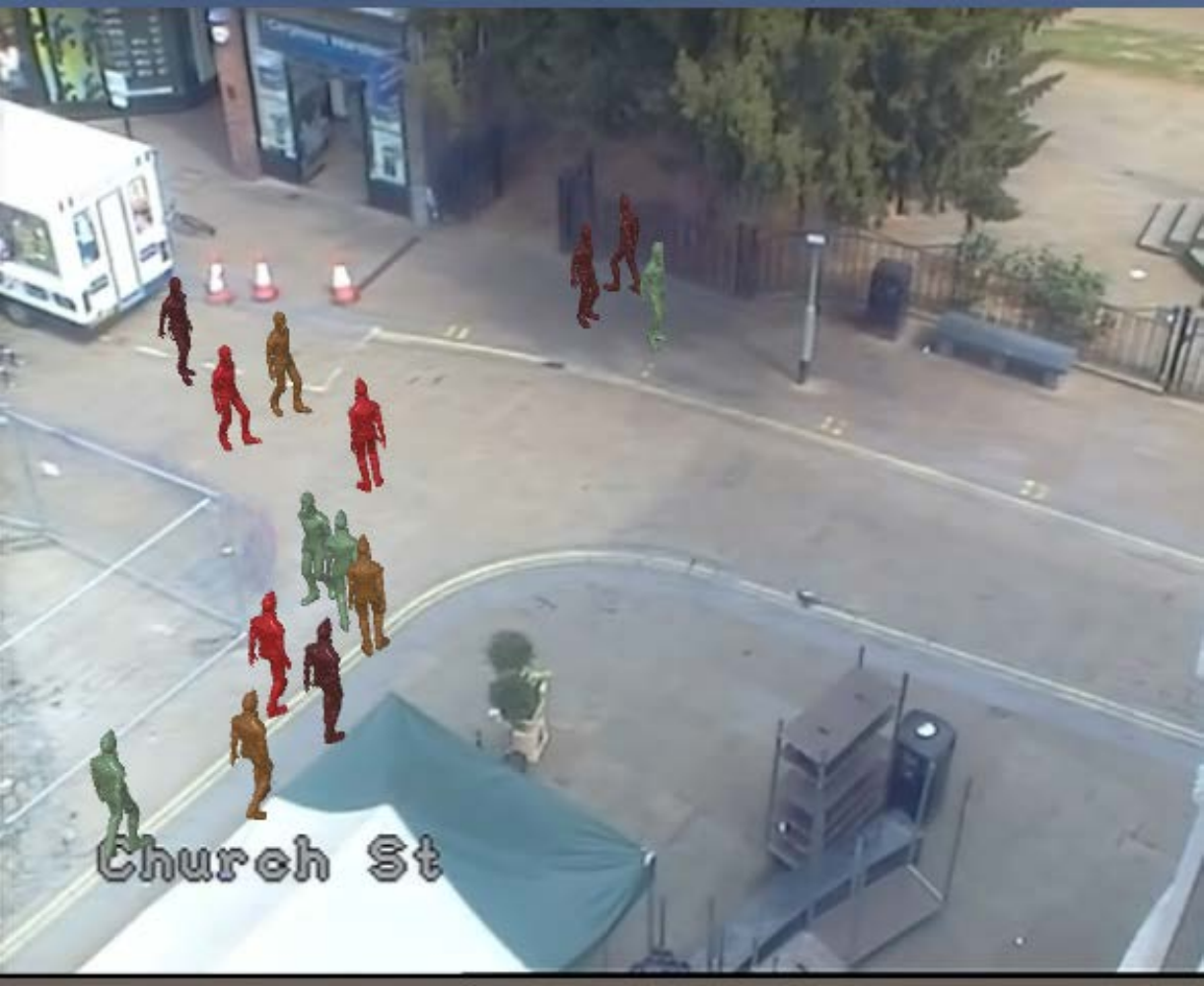}
    \includegraphics[width=0.45\columnwidth]{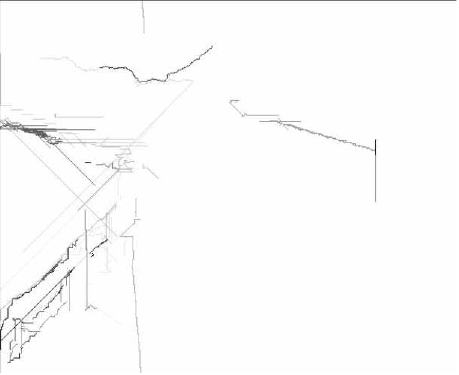}
    \caption{Side by side tracklet comparison for Road and Kvan. (a-b) Still from source Road video and tracklet. (c-d) Still from simulation and tracklet. (e-f) Still from source Kvan video and tracklet. (g-h) Still from simulation and tracklet.}
    \label{fig:5_Tracklets}
\end{figure}

\begin{figure}[!ht]
    \centering
    \includegraphics[width=0.45\columnwidth]{ROAD_SOURCE_TRACK_SIM.pdf}
    \includegraphics[width=0.45\columnwidth]{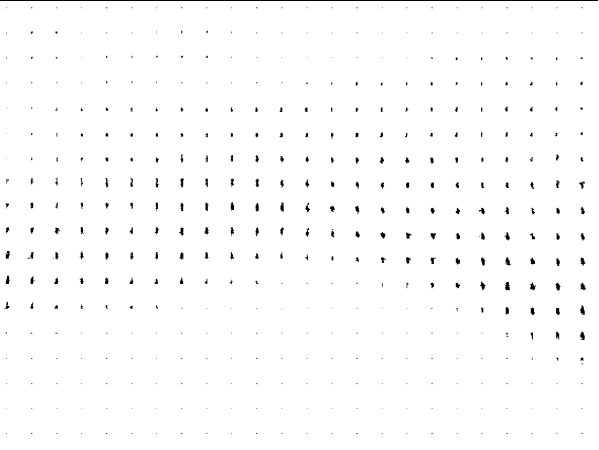}\\
    \includegraphics[width=0.45\columnwidth]{ROAD_MEDIUM_TRACK_SIM.pdf}
    \includegraphics[width=0.45\columnwidth]{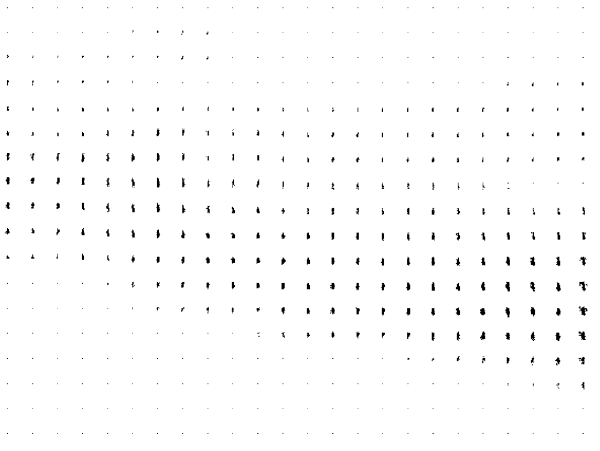}\\
    \includegraphics[width=0.45\columnwidth]{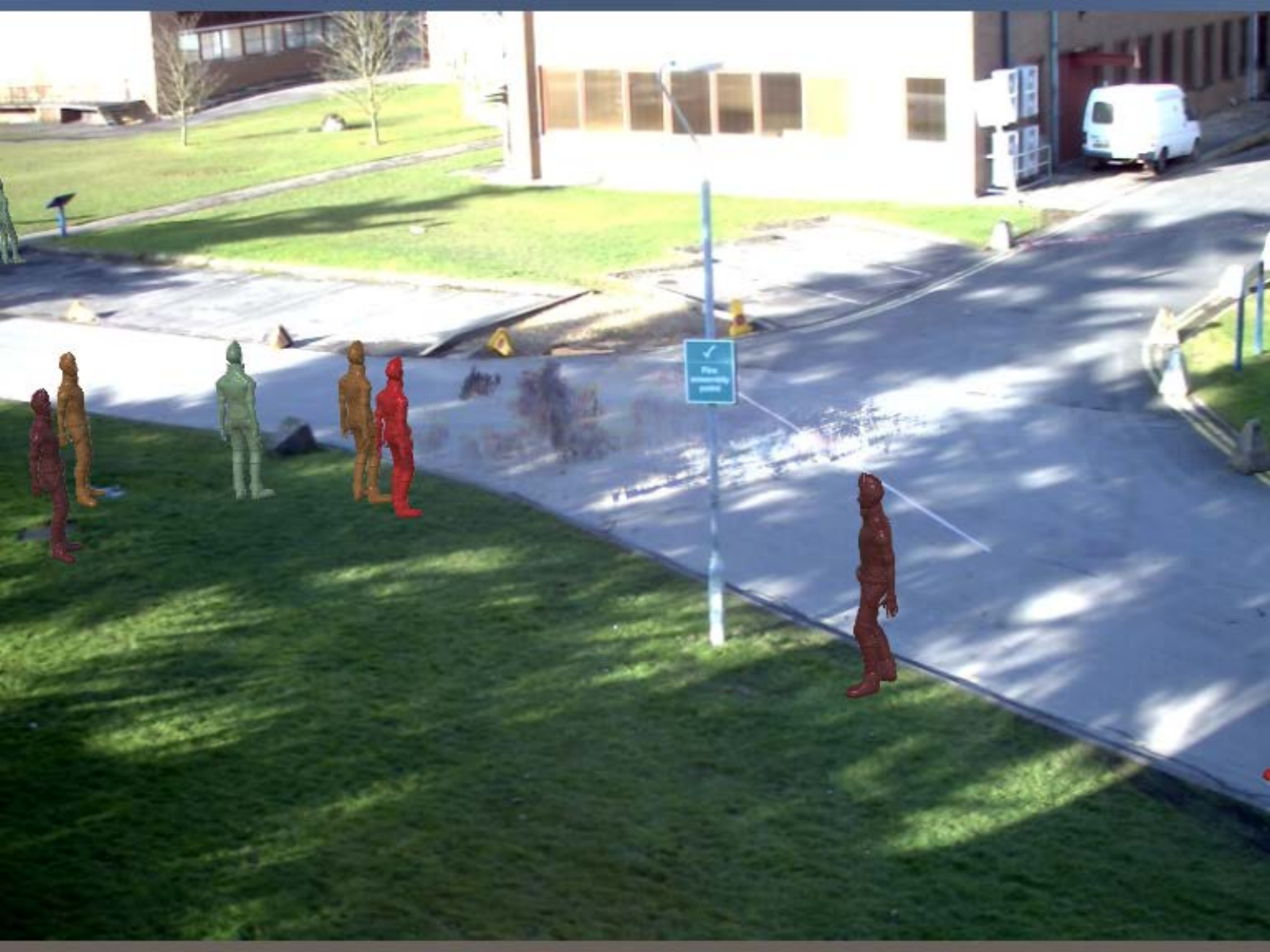}
    \includegraphics[width=0.45\columnwidth]{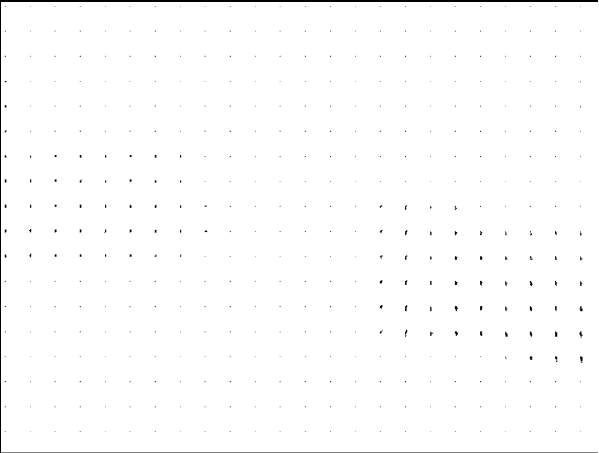}\\
    \includegraphics[width=0.45\columnwidth]{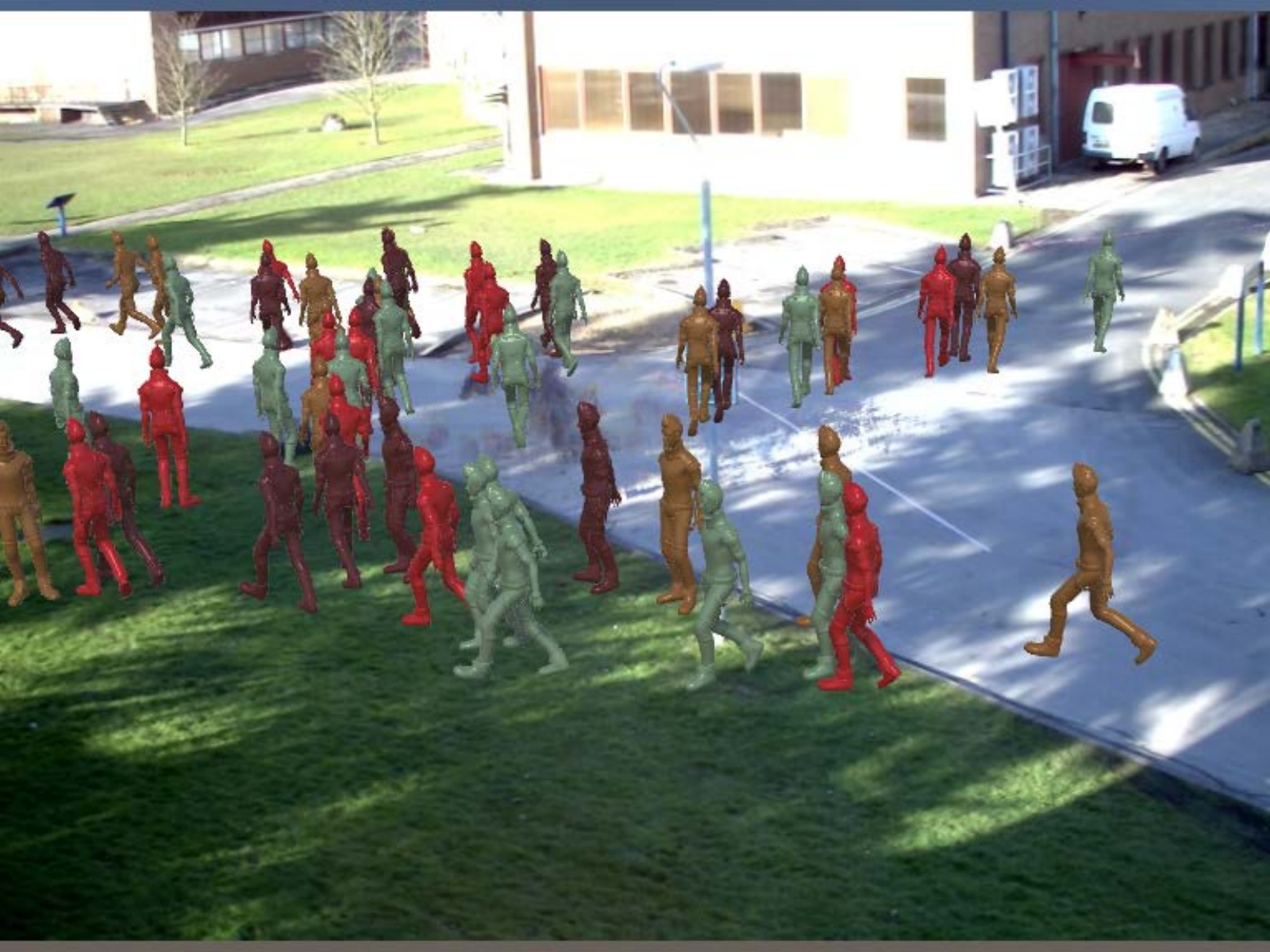}
    \includegraphics[width=0.45\columnwidth]{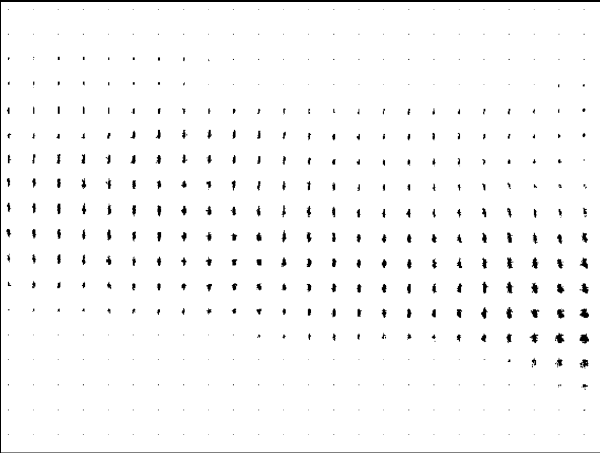}
    \caption{Histogram of Orientated Optical Flow per sequence using Road (Left example still from the video sequence and right, HOOF visualisation). (a-b) Source image, (c-d) medium, (e-f) low and (g-h) high speed and number of agent examples. }
    \label{fig:5_HOOF_S}
\end{figure}

\begin{figure}[!ht]
    \centering
    \includegraphics[width=0.45\columnwidth]{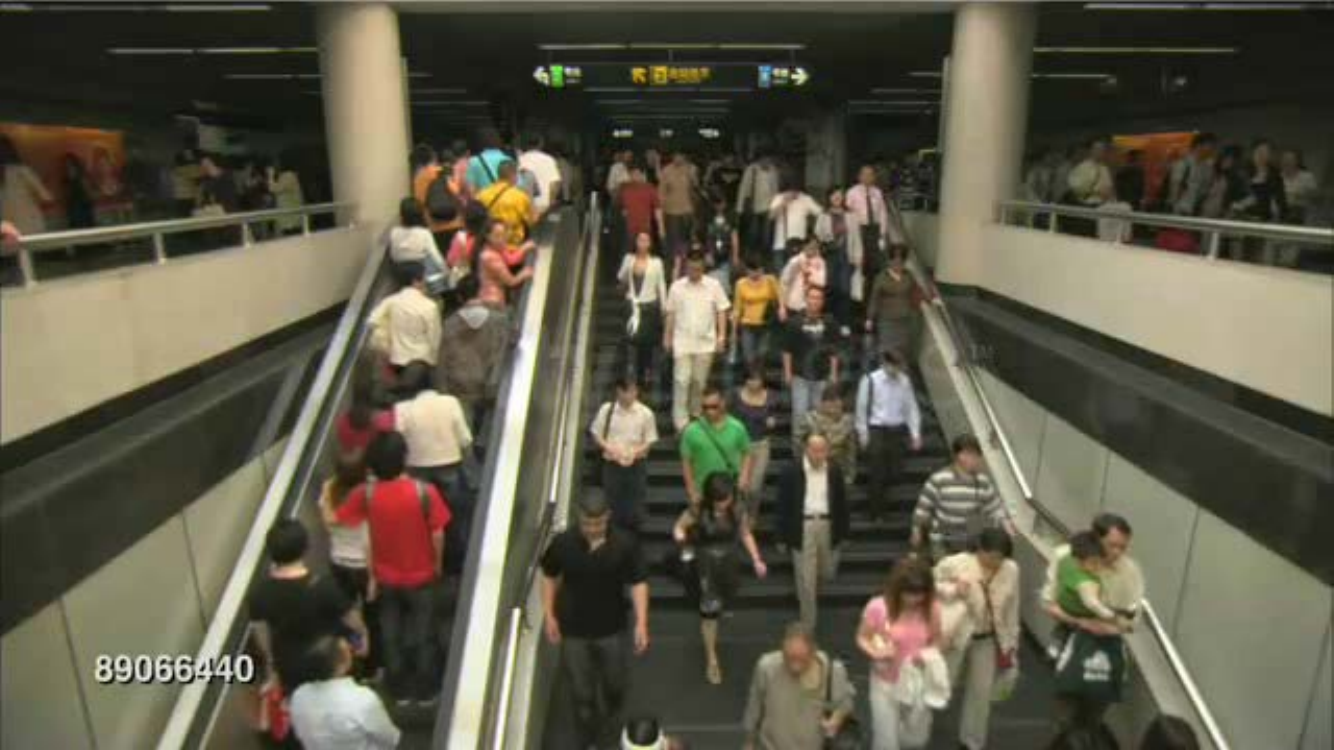}
    \includegraphics[width=0.45\columnwidth]{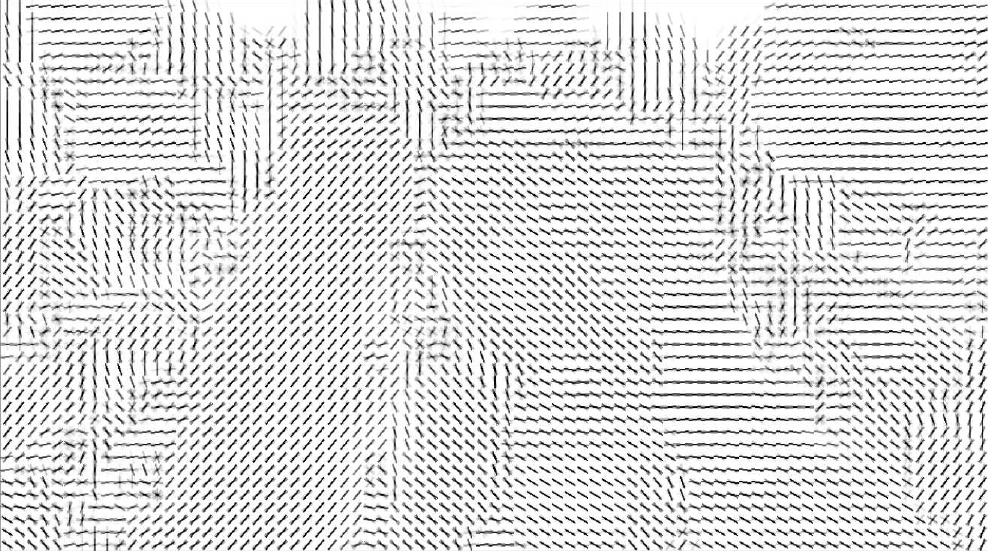}\\
    \includegraphics[width=0.45\columnwidth]{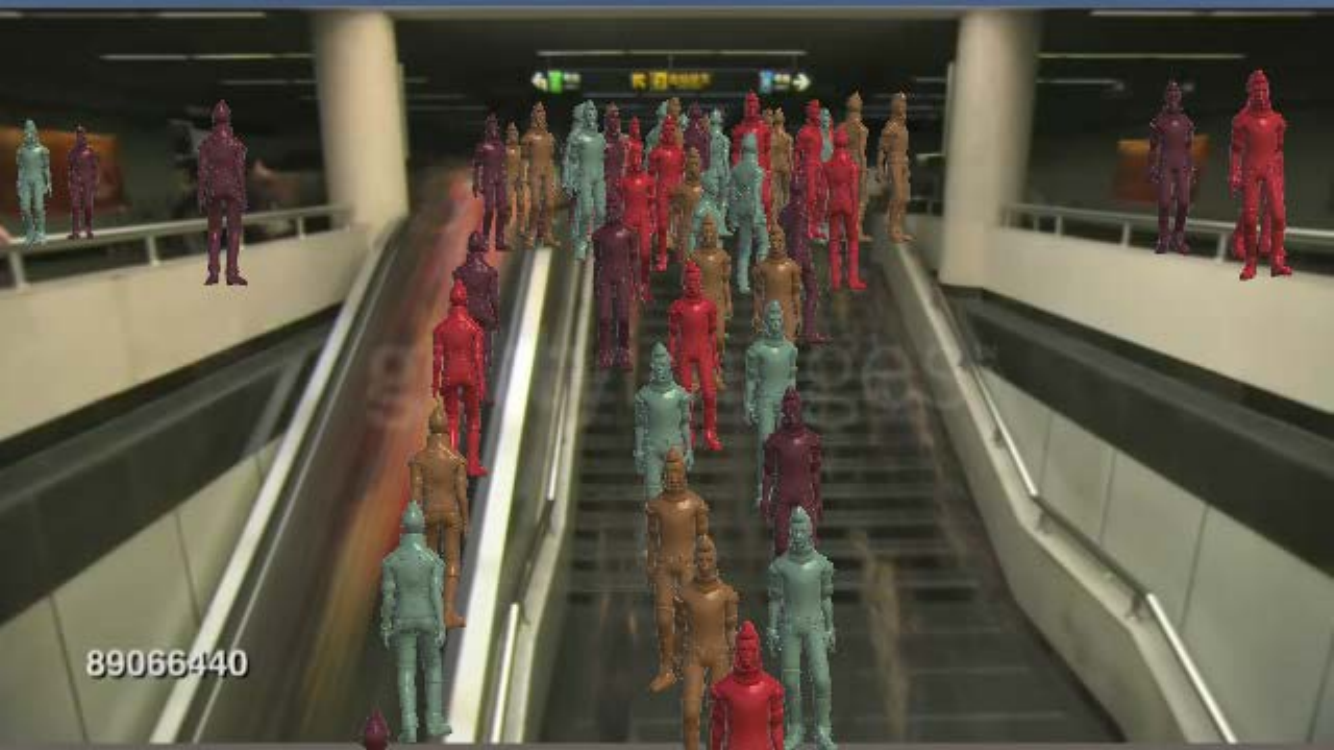}
    \includegraphics[width=0.45\columnwidth]{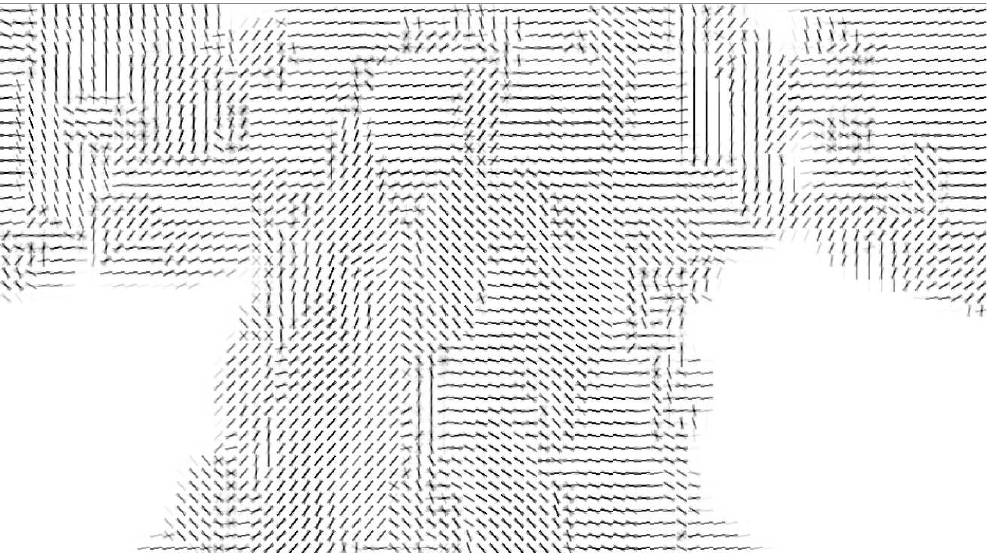}\\
    \includegraphics[width=0.45\columnwidth]{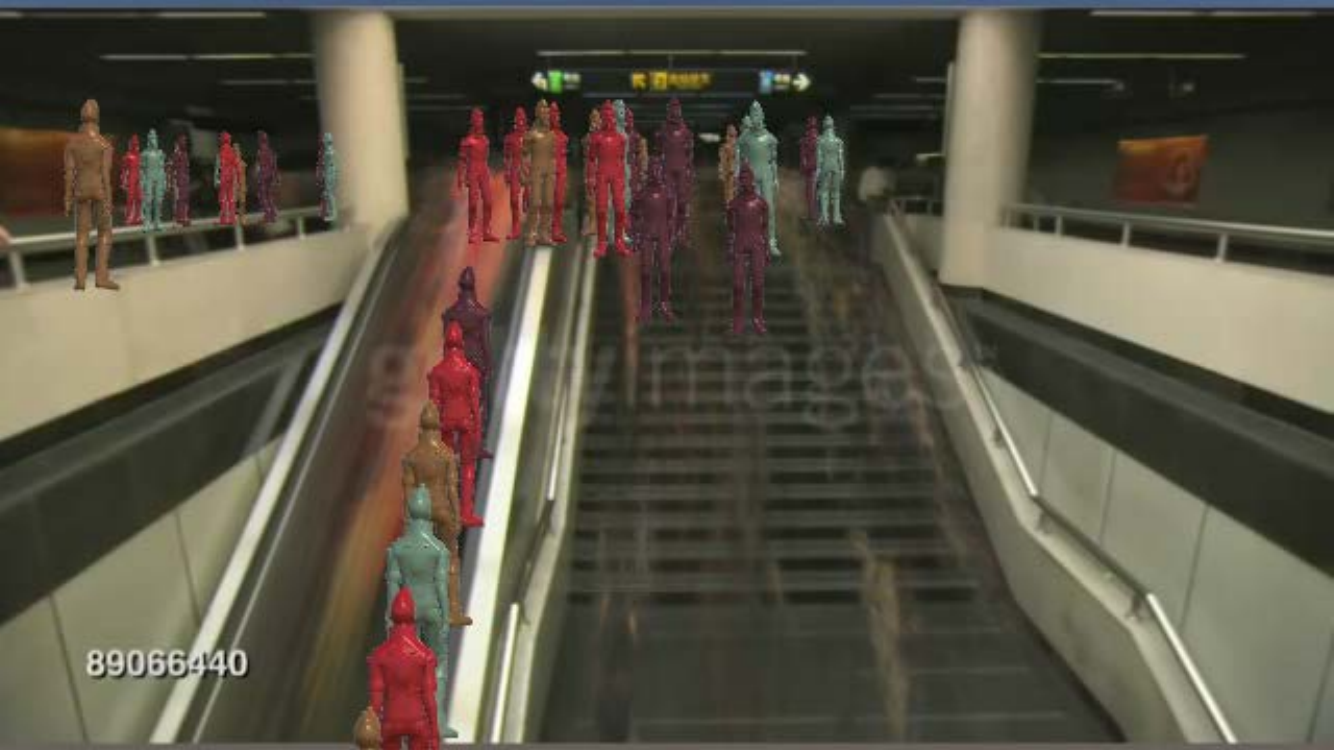}
    \includegraphics[width=0.45\columnwidth]{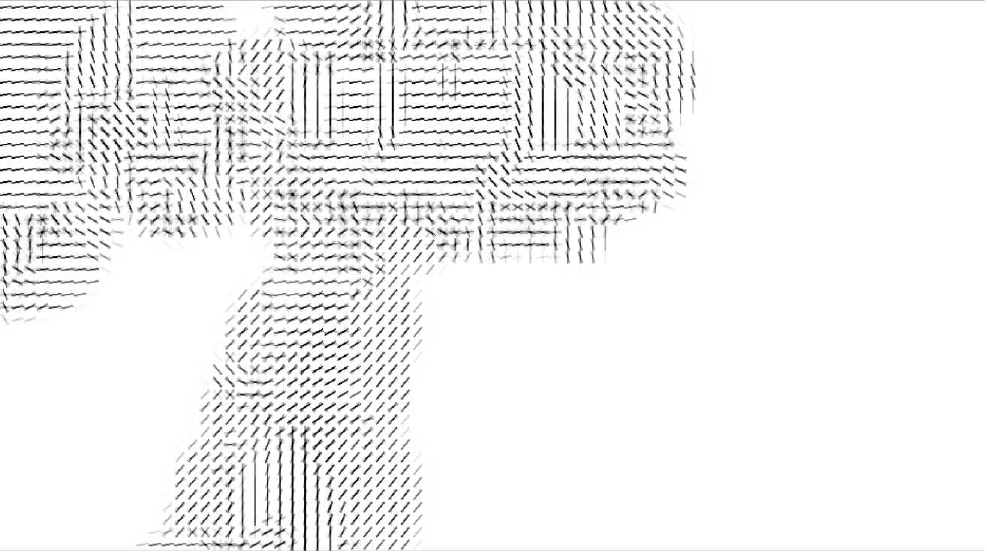}\\
    \includegraphics[width=0.45\columnwidth]{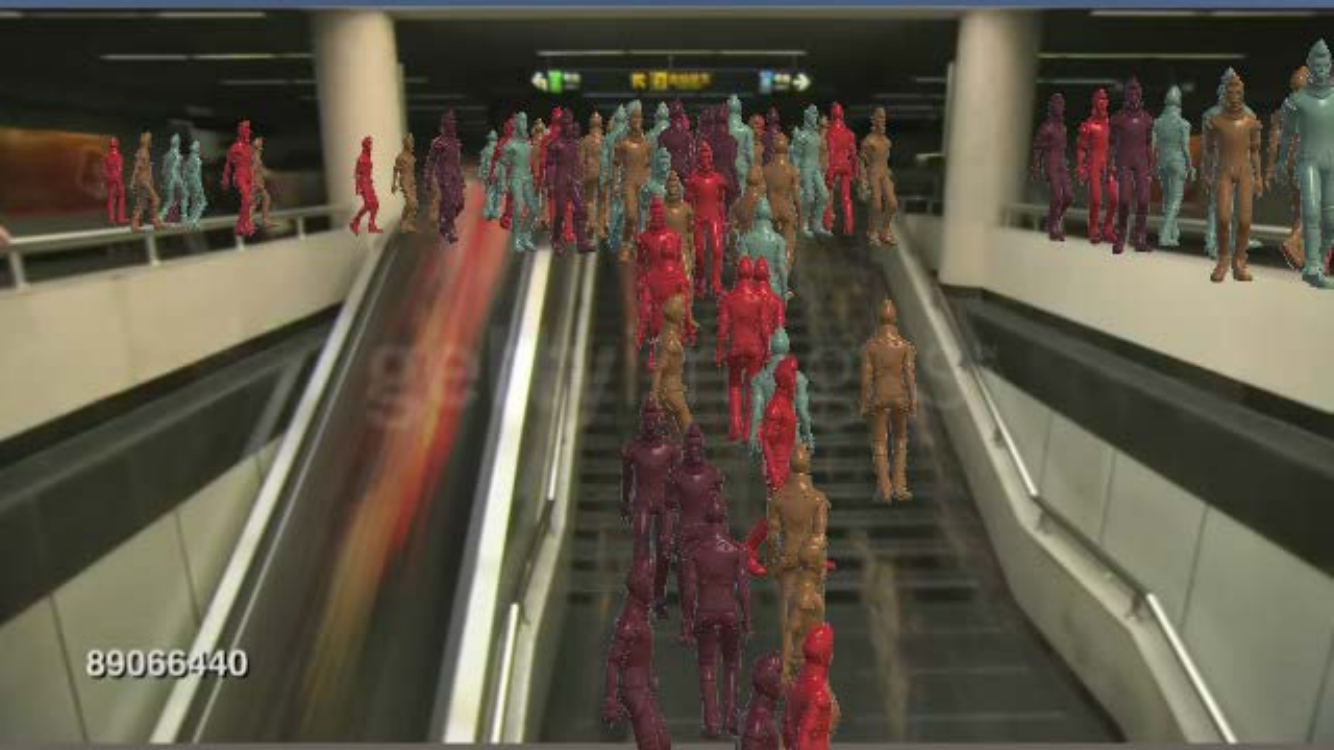}
    \includegraphics[width=0.45\columnwidth]{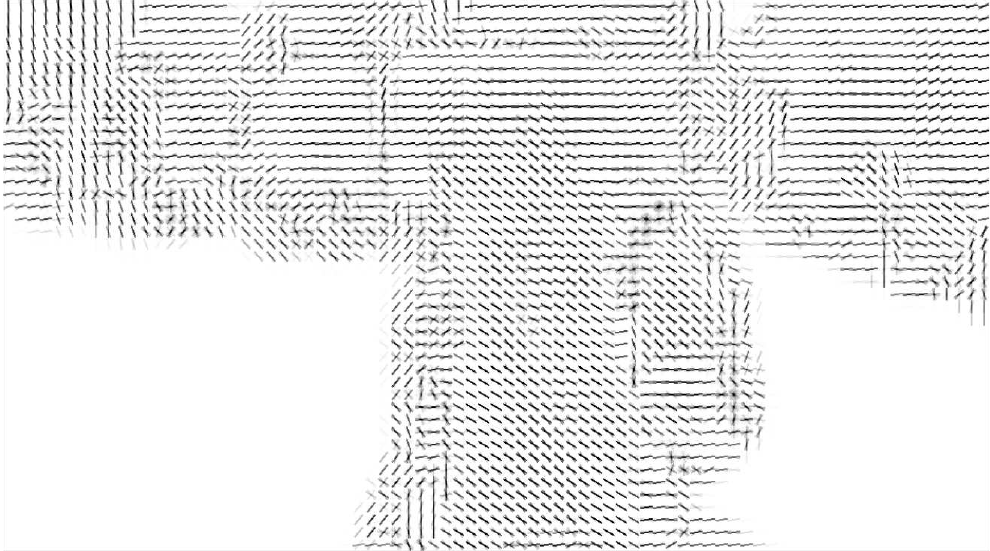}
    \caption{Histogram of Orientated Optical Flow per frame using Krad2 (Left: example still from the video sequence. Right: HOOF visualisation).(a-b) Source image, (c-d) medium, (e-f) low and (g-h) high speed and number of agent examples.}
    \label{fig:5_HOOF_F}
\end{figure}
By using the average distance from all of the three proposed features a robust system is demonstrated. However each of the individual features provides a unique insight into the simulation accuracy. For example, evaluation of the tracklets allows an insight into how accurately the simulation model replicates the movements of the source material. As such in complex scenarios where the source agents change direction a number of times, a strong dissimilarity is expected, likewise in more simplistic scenes where the simulation agents closely follow the source tracks a low dissimilarity is expected. A visual example is given in Figure \ref{fig:5_Tracklets}, where it can be seen in the first scene (a-d) that there is an obvious visual difference between the source and the simulation, whereas in the second scene (e-g) the similarity is much higher. This is visualised using the tracklet plots which represent a compound image of the tracklets over the duration of the video.

Utilising the HOOF feature per frame and per sequence, an analysis of the amount of movement and magnitude of the optical flow can be made. Visualised examples of these two features are presented in Figures \ref{fig:5_HOOF_S} \& \ref{fig:5_HOOF_F}. Figure \ref{fig:5_HOOF_S} is the compounded HOOF features for an entire sequence. The Figure \ref{fig:5_HOOF_S} (a-b) represents the source material, with (c-d) being the simulation with similar values for number of agents and their speed. Figure \ref{fig:5_HOOF_S} (e-f), (g-h) represent low and high levels of movement respectively.  Figure \ref{fig:5_HOOF_F} is the HOOF features using an individual frame. As before (a-b) is the source with (c-d), (e-f) and (g-h) being simulations with the previously mentioned parameters. In both cases its clear to see how the adjustment of speed and number of agents affects the output. Additionally effects on the tracklets can be seen. In the examples where the agent's speed is very low, parts of the scene are left unchanged by agent movement.

\section*{Conclusion}

A novel Crowd Simulation Evaluation through Composition (CSEC) framework was presented which reduces the complexity of simulation evaluation and provides tangible and relevant metrics that can be used for comparison and parametric tuning. To extract and produce simulated video, a semi-automated perspective plane extraction process is introduced which allows the conversion of source material into a composited video with controlled agents (3D models) replacing those of the humans. Through the use of a modular system, any crowd or pedestrian simulation model can be evaluated and compared by generating agent motion for use in the final visual simulation. Additionally, any video analysis feature can be utilised to evaluate similarity. Through evaluation on a large range of challenging and diverse scenes, it has been shown that the methodology presents quantifiable measures of video properties such as speed and number of agents. Utilising HVS features to replicate the human's ability to perceive movement, the framework outputs correlate well with human participant analysis of the same videos showing that the system closely emulates the Human Visual System.

The Crowd Simulation Evaluation through Composition (CSEC) framework introduces a number of key benefits over existing methods. As the framework only takes in a source and simulated video as an input, a number of the time consuming ground truth and annotation steps, such as pedestrian tracks, are reduced. The framework also allows researchers who wish to compare their algorithm against others a quick and efficient way of doing so, either by using the same well-known source material and datasets in the field or simply rerunning the framework with other pedestrian or crowd simulation algorithms to compare with. Additionally for model tuning; the proposed method can create a fast feedback loop that allows the modification of parameters to improve simulation accuracy. As the ground truth data for any simulated visualisations is already intrinsically known, and as specific testing scenarios and behaviours can be simulated, the methodology is also very suitable for the evaluation of pedestrian tracking algorithms on video data.
 
\section*{Acknowledgements}
This work is co-funded by the EU-H2020 within the MONICA project under grant agreement number 732350. The Titan X Pascal used for this research was donated by NVIDIA

\bibliographystyle{unsrt}
\bibliography{THESIS}

\begin{thebibliography}{10}

\bibitem{Asano2009}
Miho Asano, Takamasa Iryo, and Masao Kuwahara.
\newblock {A pedestrian model considering anticipatory behaviour for capacity
  evaluation}.
\newblock {\em Transportation and Traffic Theory}, 18:28, 2009.

\bibitem{Kim2012}
Sujeong Kim, Stephen~J. Guy, Dinesh Manocha, and Ming~C. Lin.
\newblock {Interactive simulation of dynamic crowd behaviors using general
  adaptation syndrome theory}.
\newblock {\em ACM SIGGRAPH Symposium on Interactive 3D Graphics and Games -
  I3D}, 1(212):55, 2012.

\bibitem{Klugl2009}
Franziska Klugl, Georg Klubertanz, and Guido Rindsfuser.
\newblock {Agent-based pedestrian simulation of train evacuation integrating
  environmental data}.
\newblock In {\em Lecture Notes in Computer Science}, volume 5803, pages
  631--638, 2009.

\bibitem{Xi2011}
Hui Xi, Seungho Lee, and Young-jun Son.
\newblock {An integrated pedestrian behavior model based on extended decision
  field theory and social force model}.
\newblock In {\em Human-in-the-Loop Simulations}, pages 69--95. 2011.

\bibitem{Portz2010}
Andrea Portz and Armin Seyfried.
\newblock {Analyzing stop-and-go waves by experiment and modeling}.
\newblock In {\em Pedestrian and Evacuation Dynamics}, pages 577--586. 2010.

\bibitem{Duives2013}
Dorine~C. Duives, Winnie Daamen, and Serge~P. Hoogendoorn.
\newblock {State-of-the-art crowd motion simulation models}.
\newblock {\em Transportation Research Part C: Emerging Technologies},
  37:193--209, 2013.

\bibitem{Charalambous2014}
P.a Charalambous, I.b Karamouzas, S.J.b Guy, and Y.a Chrysanthou.
\newblock {A data-driven framework for visual crowd analysis}.
\newblock {\em Computer Graphics Forum}, 33(7):41--50, 2014.

\bibitem{Wolinski2014}
D.~Wolinski, S.~Guy, A.H. Olivier, M.~Lin, D.~Manocha, and J.~Pettré.
\newblock Parameter estimation and comparative evaluation of crowd simulations.
\newblock {\em Computer Graphics Forum}, 2(33):303--312, 2014.

\bibitem{Guy2012}
Stephen~J Guy, Jur van~den Berg, Wenxi Liu, Rynson Lau, Ming~C Lin, and Dinesh
  Manocha.
\newblock {A statistical similarity measure for aggregate crowd dynamics}.
\newblock {\em ACM Transactions on Graphics}, 31(6):1, 2012.

\bibitem{Kapadia2011}
Mubbasir Kapadia, Matt Wang, Shawn Singh, Glenn Reinman, and Petros Faloutsos.
\newblock {Scenario space: Characterizing coverage, quality, and failure of
  steering algorithms}.
\newblock {\em Proceedings of the 2011 ACM SIGGRAPH/Eurographics Symposium on
  Computer Animation - SCA '11}, 1:53, 2011.

\bibitem{Rodreiguez2011}
Mikel Rodriguez, Josef Sivic, Ivan Laptev, and Jean-Yvesl Audibert.
\newblock {Data-driven crowd analysis in videos}.
\newblock In {\em International Conference on Computer Vision}, pages
  1235--1242, 2011.

\bibitem{Musse2012}
Soraia~R. Musse, Vinicius~J. Cassol, and Cl{\'{a}}udio~R. Jung.
\newblock {Towards a quantitative approach for comparing crowds}.
\newblock {\em Computer Animation and Virtual Worlds}, 23(1):49--57, 2012.

\bibitem{Jablonski2014}
K~Jablonski, V~Argyriou, and D~Greenhill.
\newblock {Crowd Simulation for Dynamic Environments based on Information
  Spreading and Agents' Personal Interests}.
\newblock {\em Transportation Research Procedia}, 2:412--417, 2014.

\bibitem{reynolds1987flocks}
Craig~W Reynolds.
\newblock {Flocks, herds and schools: A distributed behavioral model}.
\newblock In {\em ACM SIGGRAPH Computer Graphics}, volume~21, pages 25--34.
  ACM, 1987.

\bibitem{reynolds1999}
C.~W Reynolds.
\newblock {Steering behaviors for autonomous characters}.
\newblock In {\em Game Developers Conference}, volume 1999, pages 763--782,
  1999.

\bibitem{Helbing1998}
D~Helbing and P~Molnar.
\newblock {Self-organization phenomena in pedestrian crowds}.
\newblock {\em Condensed Matter}, pages 569--577, 1998.

\bibitem{Papadimitriou2009}
Eleonora Papadimitriou, George Yannis, and John Golias.
\newblock {A critical assessment of pedestrian behaviour models}.
\newblock {\em Transportation Research Part F: Traffic Psychology and
  Behaviour}, 12(3):242--255, 2009.

\bibitem{singh2009}
Shawn Singh, Mubbasir Kapadia, Petros Faloutsos, and Glenn Reinman.
\newblock {Steerbench: A benchmark suite for evaluating steering behaviors}.
\newblock {\em Computer Animation and Virtual Worlds}, 20(5-6):533--548, 2009.

\bibitem{Zhan2008}
Beibei Zhan, Dorothy~N. Monekosso, Paolo Remagnino, Sergio~A. Velastin, and
  Li~Qun Xu.
\newblock {Crowd analysis: A survey}.
\newblock {\em Machine Vision and Applications}, 19(5-6):345--357, 2008.

\bibitem{Pettre2009}
Julien Pettr{\'{e}}, Jan Ondrej, Anne-h{\'{e}}l{\`{e}}ne Olivier, Armel
  Cretual, and St{\'{e}}phane Donikian.
\newblock {Experiment-based modeling, simulation and validation of interactions
  between virtual walkers}.
\newblock {\em Proceedings of the 2009 ACM SIGGRAPH/Eurographics Symposium on
  Computer Animation}, 2009:189, 2009.

\bibitem{Crompton1979}
D~Crompton.
\newblock {Pedestrian delay, annoyance and risk: pre-liminary results from a 2
  years study}.
\newblock In {\em In Proceedings of PTRC Summer Annual Meeting}, pages
  275--299, 1979.

\bibitem{Wang2016}
He~Wang, Jan Ondřej, and Carol O'Sullivan.
\newblock {Path patterns: Analyzing and comparing real and simulated crowds}.
\newblock In {\em Proceedings of the 20th ACM SIGGRAPH Symposium on Interactive
  3D Graphics and Games - I3D '16}, number February, pages 49--57, 2016.

\bibitem{Wang2012a}
Qiang Wang, Yan Liu, and Juan Chen.
\newblock {Accurate indoor tracking using a mobile phone and non-overlapping
  camera sensor networks}.
\newblock {\em 2012 IEEE International Instrumentation and Measurement
  Technology Conference Proceedings}, pages 2022--2027, 2012.

\bibitem{Lerner2009}
Alon Lerner, Yiorgos Chrysanthou, Ariel Shamir, and Daniel Cohen-Or.
\newblock {Data driven evaluation of crowds}.
\newblock {\em Lecture Notes in Computer Science (including subseries Lecture
  Notes in Artificial Intelligence and Lecture Notes in Bioinformatics)}, 5884
  LNCS:75--83, 2009.

\bibitem{Lerner2010}
Alon Lerner, Yiorgos Chrysanthou, and Ariel Shamir.
\newblock {Context-dependent crowd evaluation}.
\newblock {\em Computer Graphics Forum}, 29(7):2197--2206, 2010.

\bibitem{Banerjee2011}
Bikramjit Banerjee and Landon Kraemer.
\newblock {Evaluation and comparison of multi-agent based crowd simulation
  systems}.
\newblock In {\em Agents for games and simulations II}, pages 53--66. Springer,
  2011.

\bibitem{Zanlungo2014}
Francesco Zanlungo, Dra{\v{z}}en Br{\v{s}}{\v{c}}i{\'{c}}, and Takayuki Kanda.
\newblock {Pedestrian group behaviour analysis under different density
  conditions}.
\newblock {\em Transportation Research Procedia}, 2:149--158, 2014.

\bibitem{Argyriou2007}
V.~Argyriou and T.~Vlachos.
\newblock Quad-tree motion estimation in the frequency domain using gradient
  correlation.
\newblock {\em IEEE Transactions on Multimedia}, 9(6):1147--1154, Oct 2007.

\bibitem{Argyriou2011}
V.~Argyriou.
\newblock Sub-hexagonal phase correlation for motion estimation.
\newblock {\em IEEE Transactions on Image Processing}, 20(1):110--120, Jan
  2011.

\bibitem{Bloom2015}
Victoria Bloom, Vasileios Argyriou, and Dimitrios Makris.
\newblock G3di: A gaming interaction dataset with a real time detection and
  evaluation framework.
\newblock In Lourdes Agapito, Michael~M. Bronstein, and Carsten Rother,
  editors, {\em Computer Vision - ECCV 2014 Workshops}, pages 698--712, Cham,
  2015. Springer International Publishing.

\bibitem{Bloom2014}
V.~Bloom, D.~Makris, and V.~Argyriou.
\newblock Clustered spatio-temporal manifolds for online action recognition.
\newblock In {\em 2014 22nd International Conference on Pattern Recognition},
  pages 3963--3968, Aug 2014.

\bibitem{Horn1981}
Berthold K~P Horn and Brian~G. Schunck.
\newblock {Determining optical flow}.
\newblock {\em Artificial Intelligence}, 17(1-3):185--203, 1981.

\bibitem{Lucas1981}
Bruce~D. Lucas and Takeo Kanade.
\newblock {An iterative image registration technique with an application to
  stereo vision}.
\newblock In {\em 7th International Joint Conference on Artificial
  intelligence}, volume~2, pages 674--679, 1981.

\bibitem{Sun2010}
Deqing Sun, Stefan Roth, and Michael~J. Black.
\newblock {Secrets of optical flow estimation and their principles}.
\newblock {\em Proceedings of the IEEE Computer Society Conference on Computer
  Vision and Pattern Recognition}, pages 2432--2439, 2010.

\bibitem{Sun2014}
Deqing Sun, Stefan Roth, and Michael Black.
\newblock {A quantitative analysis of current practices in optical flow
  estimation and the principles behind them}.
\newblock {\em International Journal of Computer Vision}, 106(2):115--137,
  2013.

\bibitem{Makris2002}
D~Makris and T~Ellis.
\newblock {Path detection in video surveillance}.
\newblock {\em Image and Vision Computing}, 20(12):895--903, 2002.

\bibitem{Munder2008}
Stefan Munder, Christoph Schn{\"{o}}rr, and Dariu~M Gavrila.
\newblock {Pedestrian detection and tracking using a mixture of view-based
  shape – texture models}.
\newblock {\em IEEE Transactions on Intelligent Transportation Systems},
  9(2):333--343, 2008.

\bibitem{Raptis2010}
Michalis Raptis and Stefano Soatto.
\newblock {Tracklet descriptors for action modeling and video analysis}.
\newblock {\em Lecture Notes in Computer Science (including subseries Lecture
  Notes in Artificial Intelligence and Lecture Notes in Bioinformatics)}, 6311
  LNCS(PART 1):577--590, 2010.

\bibitem{Allain2009}
Pierre Allain, Nicolas Courty, and Thomas Corpetti.
\newblock {Crowd flow characterization with optimal control theory}.
\newblock {\em Asian Conference on Computer Vision (ACCV)}, pages 279--290,
  2009.

\bibitem{hu2004Survey}
W~Hu, T~Tan, L~Wang, and S~Maybank.
\newblock {A survey on visual surveillance of object motion and behaviors}.
\newblock {\em IEEE Transactions on Systems, Man and Cybernetics, Part C},
  34(3):334--352, 2004.

\bibitem{lakoba2005}
T.~I Lakoba, D.~J Kaup, and N.~M. Finkelstein.
\newblock {Modifications of the Helbing-Moln{\'{a}}r- Farkas- Vicsek social
  force model for pedestrian evolution}.
\newblock {\em Simulation}, 81(5):339--362, 2005.

\bibitem{clarke2007}
T~L Clarke, DJ~Kaup, Linda Malone, Rex Oleson, and Mario Rosa.
\newblock {Crowd model verification using video data}.
\newblock {\em Proceedings of EMSS 2007}, pages 4--6, 2007.

\bibitem{Weber1834}
Elke~U Weber.
\newblock {De Pulsu, Resorptione, Auditu et Tactu}.
\newblock {\em Annotationes anatomicae et physiologicae}, pages 44--174, 1834.

\bibitem{Zivkovic2004}
Zoran Zivkovic.
\newblock {Improved adaptive gaussian mixture model for background
  subtraction}.
\newblock In {\em Proceedings of the 17th International Conference on Pattern
  Recognition, 2004. ICPR 2004.}, volume~2, pages 28--31, 2004.

\bibitem{Chan2008}
Antoni~B. Chan, Zhang Sheng~John Liang, and Nuno Vasconcelos.
\newblock {Privacy preserving crowd monitoring: Counting people without people
  models or tracking}.
\newblock {\em 26th IEEE Conference on Computer Vision and Pattern Recognition,
  CVPR}, 2008.

\bibitem{Karamouzas2009}
Ioannis Karamouzas, Peter Heil, Pascal {Van Beek}, and Mark~H. Overmars.
\newblock {A predictive collision avoidance model for pedestrian simulation}.
\newblock {\em Lecture Notes in Computer Science}, 5884:41--52, 2009.

\bibitem{Asano2010}
Miho Asano, Takamasa Iryo, and Masao Kuwahara.
\newblock {Microscopic pedestrian simulation model combined with a tactical
  model for route choice behaviour}.
\newblock {\em Transportation Research Part C: Emerging Technologies},
  18(6):842--855, 2010.

\bibitem{Guo2014}
Shuqiang Guo, Zhaoyang Qu, and Liqun Wang.
\newblock {Camera pose estimation using frequency analysis}.
\newblock In {\em 2014 International Conference on Information Science and
  Applications (ICISA)}, pages 3--6, 2014.

\bibitem{Zeisl2015}
Bernhard Zeisl, Torsten Sattler, and Marc Pollefeys.
\newblock {Camera pose voting for large-scale image-based localization}.
\newblock {\em Proceedings of the IEEE International Conference on Computer
  Vision}, pages 2704--2712, 2015.

\bibitem{Do2013}
Yongtae Do.
\newblock {On the neural computation of the scale factor in perspective
  transformation camera model *}.
\newblock In {\em IEEE International Conference on Control and Automation
  (ICCA)}, pages 712--714, 2013.

\bibitem{KhoongW2011}
W.~L. Khoong, W.~Y. Kow, H.~T Tan, H.~P Yoong, Kenneth Teo, and Kin Tze.
\newblock {Kalman filtering based object tracking in surveillance video
  system}.
\newblock In {\em Proceedings of the 3rd CUTSE International Conference}, 2011.

\bibitem{Hu2004}
Min Hu, Weiming Hu, and Tieniu Tan.
\newblock {Tracking people through occlusions}.
\newblock {\em Proceedings - International Conference on Pattern Recognition},
  2:724--727, 2004.

\bibitem{Wharton2008}
E.~Wharton, K.~Panetta, and S.~Agaian.
\newblock {Human visual system based similarity metrics}.
\newblock In {\em IEEE International Conference on Systems, Man and
  Cybernetics}, pages 685--690, 2008.

\bibitem{zanker1995}
Johannes~M Zanker.
\newblock {Does motion perception follow Weber's law?}
\newblock {\em Perception}, 24(4):363--372, 1995.

\bibitem{Chaudhry2009}
Rizwan Chaudhry, Avinash Ravichandran, Gregory Hager, and Ren?? Vidal.
\newblock {Histograms of oriented optical flow and Binet-Cauchy kernels on
  nonlinear dynamical systems for the recognition of human actions}.
\newblock {\em 2009 IEEE Computer Society Conference on Computer Vision and
  Pattern Recognition Workshops, CVPR Workshops 2009}, pages 1932--1939, 2009.

\bibitem{Loy2013}
Chen~Change Loy, Shaogang Gong, and Tao Xiang.
\newblock {From semi-supervised to transfer counting of crowds}.
\newblock {\em 2013 IEEE International Conference on Computer Vision}, pages
  2256--2263, 2013.

\bibitem{PETS2009}
J.~Ferryman and A.~Ellis.
\newblock {PETS2010: Dataset and challenge}.
\newblock {\em Proceedings - IEEE International Conference on Advanced Video
  and Signal Based Surveillance, AVSS 2010}, pages 143--148, 2010.

\bibitem{Simonnet2011}
Damien Simonnet, Sergio~A. Velastin, James Orwell, and Esin Turkbeyler.
\newblock {Selecting and evaluating data for training a pedestrian detector for
  crowded conditions}.
\newblock {\em 2011 IEEE International Conference on Signal and Image
  Processing Applications, ICSIPA 2011}, pages 174--179, 2011.

\bibitem{bhattachayya1943}
A~Bhattachayya.
\newblock {On a measure of divergence between two statistical population
  defined by their population distributions}.
\newblock {\em Bulletin Calcutta Mathematical Society}, 35:99--109, 1943.

\end{thebibliography}

\end{document}